\pdfoutput=1

\documentclass[11pt]{article}

\usepackage[final]{acl}

\usepackage{times}
\usepackage{latexsym}
\usepackage{mdframed}
\usepackage{enumitem}
\usepackage[T1]{fontenc}

\usepackage[utf8]{inputenc}

\usepackage{microtype}

\usepackage{inconsolata}
\usepackage{longtable}
\usepackage{booktabs}
\usepackage{graphicx}
\usepackage{xspace}

\usepackage{multirow}
\usepackage{amssymb}
\usepackage{makecell}
\usepackage{caption}
\usepackage{subcaption}
\usepackage{wrapfig}
\usepackage{tabularx}
\usepackage{array}

\newcommand{\ie}{\textit{i.e., }}
\newcommand{\eg}{\textit{e.g., }}

\newcommand{\method}{\textsc{EgoIllusion}\xspace}

\title{
    \begin{center}
        \begin{tabular}{@{}c@{\hskip 0.2em}c@{}}
            \raisebox{-0.3\height}{\includegraphics[height=2.6em]{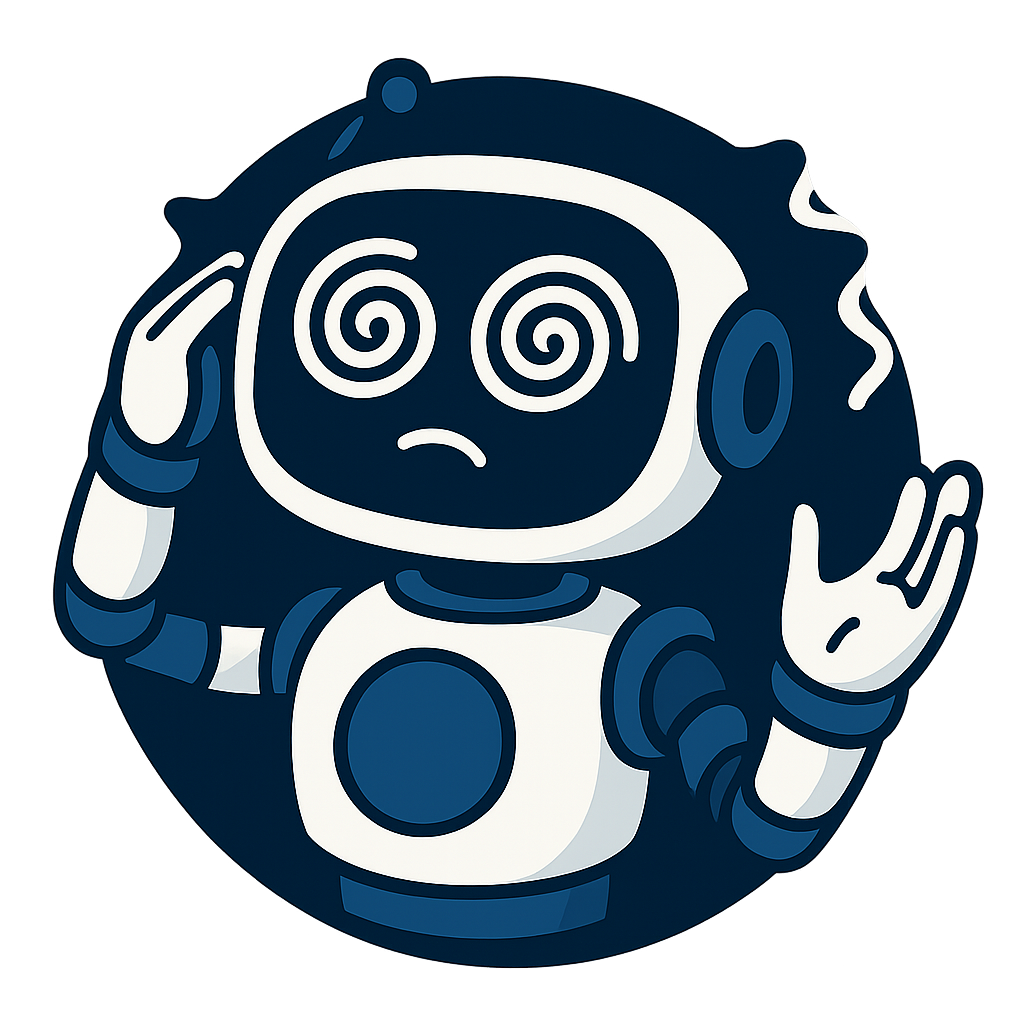}} &
            \makebox[0.8\textwidth][c]{%
                \begin{tabular}{c}
                    \textbf{\LARGE \method: Benchmarking Hallucinations} \\
                    \textbf{\LARGE in Egocentric Video Understanding}
                \end{tabular}
            }
        \end{tabular}
    \end{center}
}

\author{
    Ashish Seth$^{1*}$, Utkarsh Tyagi$^{1*}$, Ramaneswaran Selvakumar$^{1}$, Nishit Anand$^{1}$ \\ \bf Sonal Kumar$^{1}$ \bf Sreyan Ghosh$^{1}$, \bf Ramani Duraiswami$^{1}$, \bf Chirag Agarwal$^{2}$, \bf Dinesh Manocha$^{1}$
    \\$^{1}$University of Maryland, College Park, $^{2}$University of Virginia\\
    Correspondence: \texttt{aseth125@umd.edu}
}

\begin{document}
\maketitle
\begin{abstract}

\looseness=-1 Multimodal Large Language Models (MLLMs) have demonstrated remarkable performance in complex multimodal tasks. While MLLMs excel at visual perception and reasoning in third-person and egocentric videos, they are prone to hallucinations, generating coherent yet inaccurate responses. We present \method, a first benchmark to evaluate MLLM hallucinations in egocentric videos. \method comprises 1,400 videos paired with 8,000 human-annotated open and closed-ended questions designed to trigger hallucinations in both visual and auditory cues in egocentric videos. Evaluations across ten MLLMs reveal significant challenges, including powerful models like GPT-4o and Gemini, achieving only~59\% accuracy. \method lays the foundation in developing robust benchmarks to evaluate the effectiveness of MLLMs and spurs the development of better egocentric MLLMs with reduced hallucination rates. Our benchmark will be open-sourced for reproducibility\footnote{* Equal Contribution, Please find the benchmark \href{https://sites.google.com/view/egoillusion-demo/home}{here}}.
\end{abstract}

\section{Introduction}
\label{sec:intro}
\looseness=-1 Recent advances in Multimodal Large Language Models (MLLMs) have expanded their capabilities beyond image understanding to video comprehension, enabling advanced multimodal perception and reasoning~\citep{achiam2023gpt, dubey2024llama, ye2024mplug, wang2024qwen2, wu2024deepseekvl2mixtureofexpertsvisionlanguagemodels}. Depending on the camera viewpoint and observer’s position, videos can be categorized as third-person (\textit{exocentric}) videos, captured from a stationary or spectator perspective, and first-person (\textit{egocentric}) videos, recorded from an active observer’s viewpoint~\citep{Jia_2024_CVPR, luo2024shoesliftingegocentricperspective, grauman2024ego}. Egocentric videos captured from wearable devices primarily capture human-object interactions, providing rich multi-sensory information, including actions performed, object appearances, and the sounds produced during interactions~\citep{chen2024soundingactionslearningactionssound, grauman2024ego, kim2024palmpredictingactionslanguage, hatano2024multimodalcrossdomainfewshotlearning, chen2024action2soundambientawaregenerationaction}. Unlike exocentric videos, where objects often remain static, egocentric interactions dynamically alter object states (\eg opening a bottle or turning on a device), making inference of object properties and their temporal evolution more challenging.

\begin{figure}
    \centering
    \includegraphics[width=0.9\linewidth]{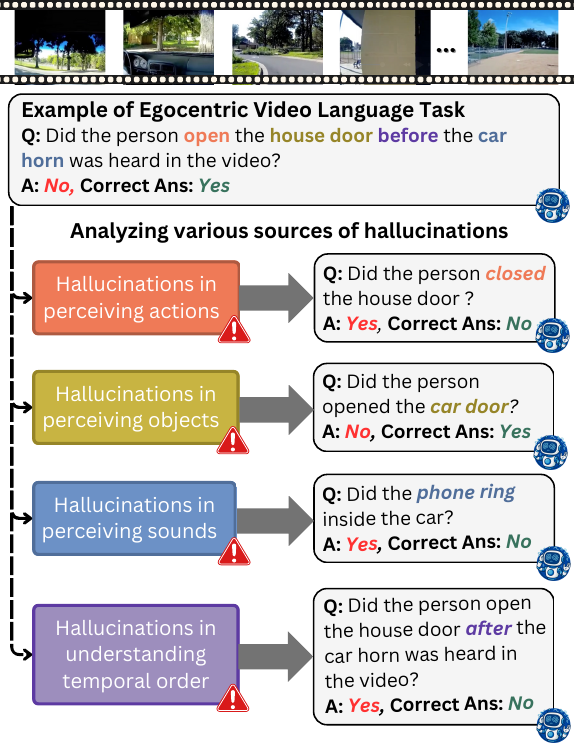}
    \caption{\small Illustration of various sources of hallucination encountered by MLLMs, such as Gemini~\cite{team2024gemini}, while performing an egocentric video-language task involving temporal reasoning between two distinct events, such as a \textit{person opening a house door} and \textit{a car horn is heard}.}
    \label{fig:hero_diag_ego}
\end{figure}

\looseness=-1 Although MLLMs demonstrate strong performance on standard image and video benchmarks~\citep{fu2024video}, they remain susceptible to hallucinations, producing coherent but incorrect interpretations of sensory input that diverge from reality. As illustrated in Fig.~\ref{fig:hero_diag_ego}, state-of-the-art MLLMs such as Gemini~\citep{team2024gemini} exhibit a high rate of hallucination when processing multisensory information in egocentric video, such as human actions, visual objects, and ambient sounds. Accurate perception of such elements is critical in performing common egocentric video-language tasks, including temporal reasoning between events.  

\looseness=-1\noindent{\textbf{\method vs. Existing Benchmarks.}} As shown in Table~\ref{tab:benchmark_comparison}, we compare \method with existing hallucination benchmarks. Prior work has primarily focused on hallucinations in \textit{static visual attributes} like object properties \citep{grauman2022ego4d, kaul2024throne, wang2023llm} or factual inconsistencies \citep{wang2024videohallucer, guan2024hallusionbench}, with \textit{limited attention to video-based hallucinations}. While VideoHallucer \citep{wang2024videohallucer} targets \textit{exocentric} videos, it overlooks the unique challenges of egocentric settings, such as occlusions from hand movements, action-centric narratives prone to temporal hallucinations \citep{grauman2022ego4d}, and rich multisensory cues such as auditory cues, often misaligned by MLLMs~\citep{su2024icml}.

\begin{table}[t]
\centering
\renewcommand{\arraystretch}{1.125}
\resizebox{\columnwidth}{!}{
\begin{tabular}{l|c|cc|rlrl}
\toprule
\multirow{2}{*}{\textbf{Benchmark}} & \multirow{2}{*}{\textbf{Size}} & \multicolumn{2}{c|}{\textbf{Modality}} & \multicolumn{4}{c}{\textbf{Skills}} \\ \cline{3-8}
  &                                & \textbf{Vision}    & \textbf{Audio}    & \multicolumn{2}{c}{\textbf{Perception}}  & \multicolumn{2}{c}{\textbf{Reasoning}}   \\ \midrule
POPE~\citep{li2023evaluating}  & 3k  & \textcolor{green}{\checkmark}  & \textcolor{red}{$\times$}  & 3k  & \textcolor{green}{\checkmark}  & 0  & \textcolor{red}{$\times$}  \\
HallusionBench~\citep{guan2024hallusionbench}  & 1.1k  & \textcolor{green}{\checkmark}  & \textcolor{red}{$\times$}  & 0  & \textcolor{red}{$\times$}  & 1.1k  & \textcolor{green}{\checkmark}  \\
MMHal-Bench~\citep{sun2023aligning}  & 0.1k  & \textcolor{green}{\checkmark}  & \textcolor{red}{$\times$}  & 0.05k  & \textcolor{green}{\checkmark}  & 0.05k  & \textcolor{green}{\checkmark}  \\
Bingo~\citep{cui2023holistic}  & 0.4k  & \textcolor{green}{\checkmark}  & \textcolor{red}{$\times$}  & 0  & \textcolor{red}{$\times$}  & 0.4k  & \textcolor{green}{\checkmark}  \\
EasyDetect~\citep{chen2024unified}  & 0.4k  & \textcolor{green}{\checkmark}  & \textcolor{red}{$\times$}  & 0.4k  & \textcolor{green}{\checkmark}  & 0  & \textcolor{red}{$\times$}  \\
VHTest~\citep{huang2024visual}  & 1.2k  & \textcolor{green}{\checkmark}  & \textcolor{red}{$\times$}  & 0.6K  & \textcolor{green}{\checkmark}  & 0.6K  &\textcolor{green}{\checkmark}   \\
VALOR~\citep{chen2023valor}  & 0.2k  & \textcolor{green}{\checkmark}  & \textcolor{red}{$\times$}  & 0.2k  & \textcolor{green}{\checkmark}  & 0  & \textcolor{red}{$\times$}  \\
VideoHallucer~\citep{wang2024videohallucer}  & 1.8k  & \textcolor{green}{\checkmark}  & \textcolor{red}{$\times$}  & 0.9k  & \textcolor{green}{\checkmark}  & 0.9k & \textcolor{green}{\checkmark}  \\ \midrule
\textbf{\method} \textit{(ours)}  & \textbf{8k}  & \textcolor{green}{\checkmark}  & \textcolor{green}{\checkmark}  & \textbf{4.0k}  & \textcolor{green}{\checkmark}  & \textbf{4.0k}  & \textcolor{green}{\checkmark}  \\
\bottomrule
\end{tabular}}
\caption{\small Comparison of \method with existing multimodal hallucination benchmarks. \method covers both vision and audio modality, while having the highest number of perception and reasoning-based questions.}
\label{tab:benchmark_comparison}
\end{table}

{\noindent \textbf{Main Contributions.}} In this work, we introduce \method, a benchmark designed to evaluate hallucinations in MLLMs when processing egocentric videos. \method includes over \textit{1,400} egocentric videos, ranging from 30 seconds to 5 minutes, along with \textit{8,000} human-annotated question-answer pairs. These questions assess hallucinations across diverse egocentric video-language tasks that demand advanced multimodal perception and reasoning skills. To examine hallucinations in multimodal perception, we design tasks with intricate question-answer pairs that test MLLMs' ability to infer multisensory information accurately. These tasks require models to reason about actions, sounds, and visual objects involved in human-object interactions recorded from a first-person perspective. To this end, we develop novel egocentric video-language tasks to reliably evaluate MLLMs' temporal reasoning by integrating diverse sensory cues. Additionally, we introduce hallucination questions focused on contextual and causal reasoning, which require models to infer the presence or absence of human actions, sounds, and objects before generating factually grounded responses. Our key contributions are:

\begin{itemize}[leftmargin=*, noitemsep, topsep=0pt]
    \item We present \method, the first hallucination benchmark specifically designed for egocentric video. \method features \textit{8,000} question-answer pairs that capture diverse human-object interactions and enable a systematic evaluation of hallucinations across multimodal perception and understanding.  
    \item We evaluate 10 MLLMs, including eight open-source and two proprietary models, demonstrating that state-of-the-art MLLMs exhibit a high degree of hallucinations, with the best performance of only 59\% on \method.  
    \item We perform extensive analysis on the models' responses and uncover key insights such as skill-wise hallucinations, challenges MLLMs face in attending multisensory input, and hallucination against diverse egocentric video-language tasks.     
\end{itemize}

\section{Related works}
\begin{figure*}
    \centering
    \includegraphics[width=1.0\linewidth]{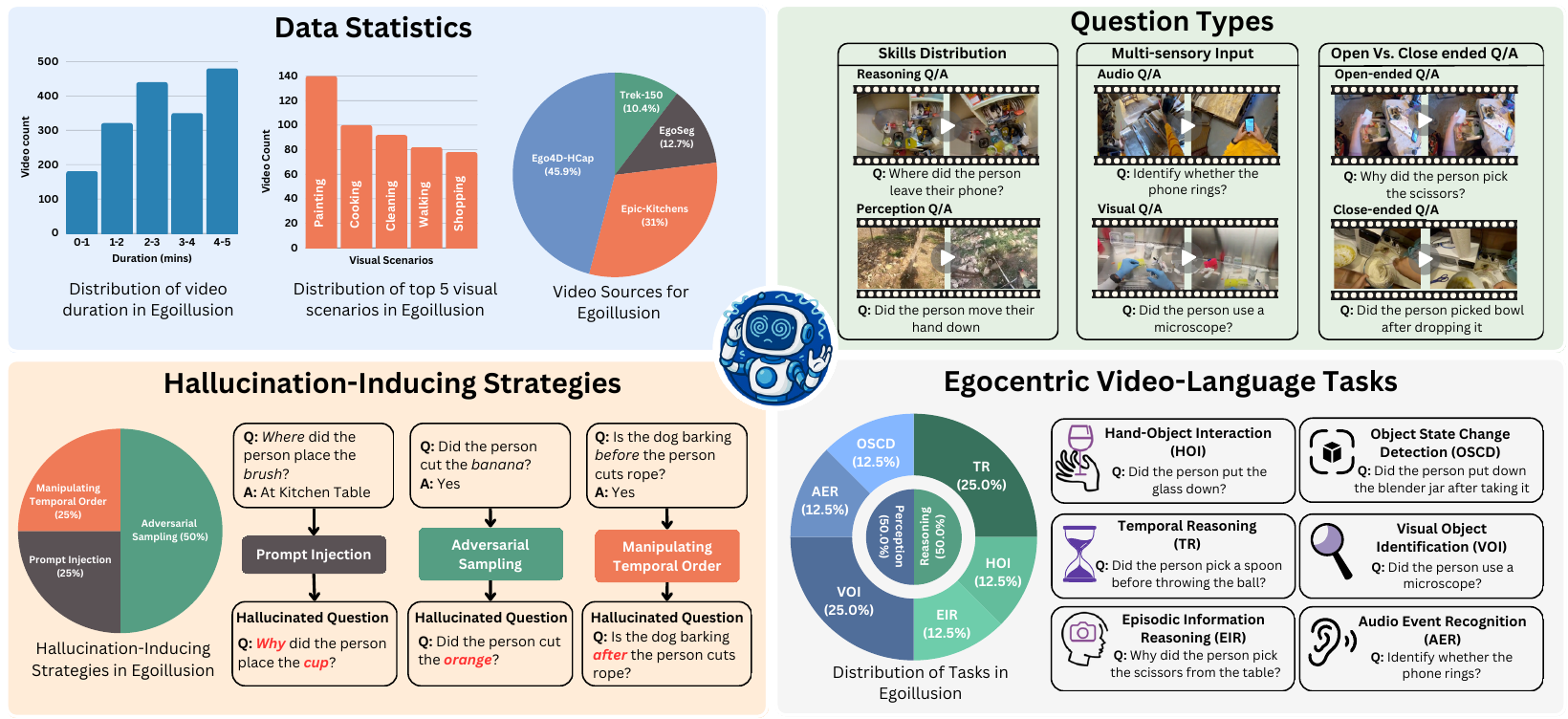}
    \caption{\small Overview of the \method benchmark. \method is the first hallucination benchmark for egocentric videos, featuring 8,000 human-annotated questions covering diverse egocentric video-language tasks. It presents three core challenges: (1) \textbf{Perception vs Reasoning:} distinguishing between perceptual and reasoning skills by evaluating object recognition, action understanding, and scene inference; (2) \textbf{Multisensory Inputs:} integrating visual and auditory cues, such as object appearance, human actions, and environmental sounds, to assess multimodal alignment; (3) \textbf{Question Types:} supporting both closed-ended and open-ended questions, requiring models to answer factually grounded queries while reasoning about events and interactions.}
    \label{fig:main_diag}
\end{figure*}
\label{sec:related}
\noindent{\textbf{Egocentric Video Understanding.}} Egocentric video understanding has gained momentum with benchmarks like Ego4D~\citep{grauman2022ego4d}, Ego-Exo4D~\citep{grauman2024ego}, and EPIC-KITCHENS100~\citep{Damen2022RESCALING}, which offer large-scale, annotated recordings for tasks such as activity recognition and object interaction. Multimodal datasets like QaEgo4D~\citep{barmann2022did} and EgoSchema~\citep{mangalam2023egoschema} further enrich semantic understanding by incorporating language. Recent modeling efforts—GroundVQA~\citep{di2024grounded}, Encode-Store-Retrieve~\citep{shen2024encode}, and R-VLM~\citep{xu2023retrieval}—focus on long-horizon reasoning and factual consistency. However, existing benchmarks largely emphasize factual recall and recognition, lacking a systematic evaluation of hallucination. \textit{Our work fills this gap by introducing the first benchmark designed to assess hallucination in egocentric video understanding.}

\noindent{\textbf{Multimodal Large Language Models.}} 
Recent advances in MLLMs have extended their capabilities beyond static image understanding to complex video-based perception and reasoning, incorporating both visual and auditory signals~\citep{wang2024qwen2, li2024llavanextinterleavetacklingmultiimagevideo, li2024llavaonevisioneasyvisualtask, han2023imagebindllmmultimodalityinstructiontuning}. While some models rely solely on visual inputs, others explicitly integrate audio to enrich multimodal understanding~\citep{MiniCPM2024, cheng2024videollama}. Most are trained primarily on third-person videos; only a few incorporate egocentric data. For instance, MiniCPM~\citep{MiniCPM2024} uses only third-person videos, VideoLLaMA 2 and 3~\citep{cheng2024videollama, zhang2025videollama} mix third-person and egocentric views, and MMEgo~\citep{ye2024mm} focuses exclusively on egocentric content. Despite strong performance on standard benchmarks~\citep{fu2024video, li2024mvbenchcomprehensivemultimodalvideo}, \textit{we find that these models remain susceptible to hallucinations, with the best achieving just 59\% accuracy on \method.}

\section{The \method Benchmark}
\begin{figure*}[t]
    \centering
    \includegraphics[width=1.0\textwidth]{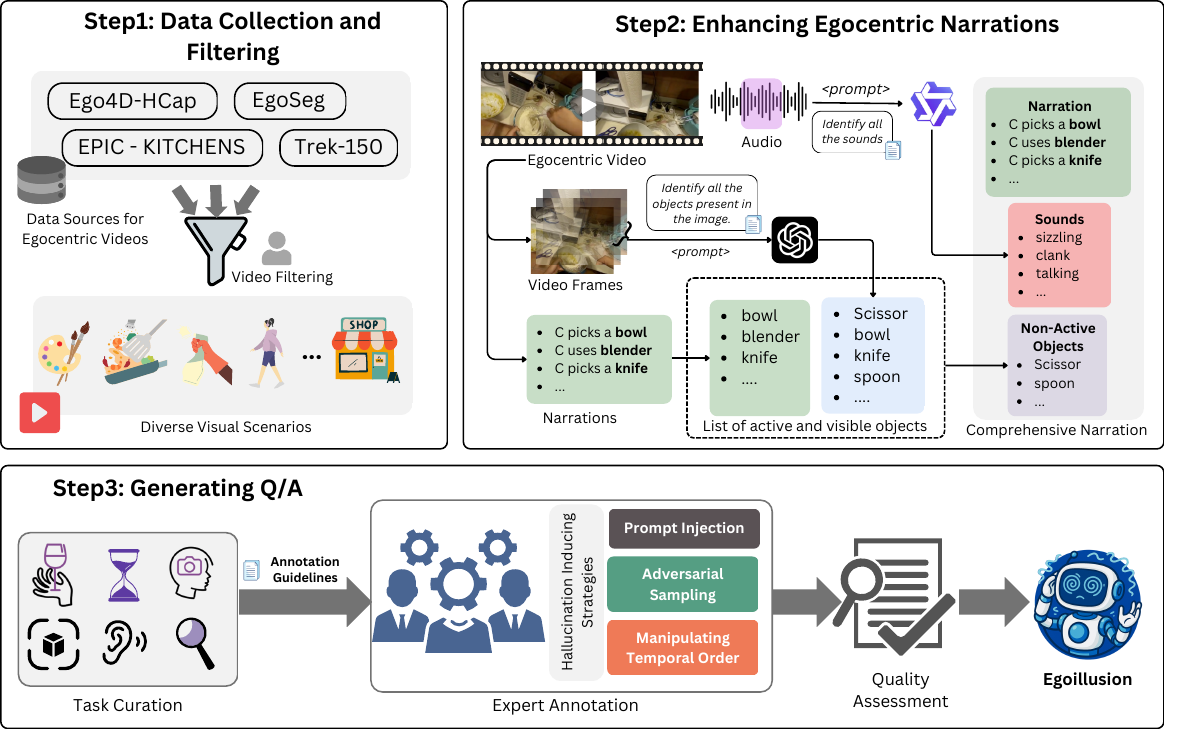}
    \caption{\small Illustration of the \method data construction pipeline. We first collect egocentric videos with detailed narrations from open-source datasets like Ego4D-HCap~\citep{islam2024video} and EPIC-KITCHENS~\citep{Damen2022RESCALING}, and manually filter them to ensure diverse visual scenarios (e.g., cooking, painting). We then develop an automated pipeline to enhance narrations by inferring active/inactive object states using GPT-4o~\citep{achiam2023gpt} and incorporating environmental sounds via Qwen2-Audio~\citep{chu2024qwen2}. Finally, we generate question-answer pairs through a rigorous human annotation process involving egocentric task design, guideline creation for inter-annotator consistency, applying hallucination-inducing strategies, and QA review.} 
    \label{fig:comp_narr}
\end{figure*}
\subsection{Overview}
We introduce \method, a novel benchmark to systematically evaluate hallucination in MLLMs across a diverse set of egocentric video-language tasks. \method consists of egocentric videos spanning various visual scenarios (Fig.~\ref{fig:main_diag}), including question types requiring perceptual and reasoning skills. The benchmark features questions based on multi-sensory inputs, including visual and auditory modalities and open- and closed-ended formats. Additionally, it incorporates a range of hallucination-inducing strategies from various egocentric video-language tasks. Below, we describe the data construction pipeline of \method.
\subsection{Data Collection and Filtering}
We illustrate our data construction pipeline in Fig.\ref{fig:comp_narr}. The videos included in \method are carefully selected from a diverse collection of egocentric datasets including Ego4D-HCap~\citep{islam2024video}, EgoSeg~\citep{poleg_wacv16_compactcnn}, EPIC-KITCHENS~\citep{Damen2022RESCALING} and Trek-150~\citep{TREK150ijcv}, covering a wide range of visual scenarios such as meal preparation in a kitchen, painting a canvas, assembling furniture and navigating urban environments (additional details on these can be found in Appendix~\ref{sec:data_sources}). The videos in \method span a broad range of durations, from short clips of 30 seconds to extended recordings exceeding 5 minutes.

\looseness=-1 To ensure coverage of diverse visual content and meaningful temporal dynamics, the dataset construction of the \method includes a manual filtering step, which involves selecting videos that depict varied object interactions and human activities. For instance, a video showing a person transitioning from preparing ingredients to cooking and serving a meal is retained, but videos with minimal variation, such as someone stirring a pot for several minutes or walking down an empty hallway without significant interaction, are excluded. This filtering ensures that the dataset emphasises visually and temporally rich scenarios crucial for generating complex queries and effectively evaluating hallucination in egocentric video-language models.

\subsection{Enhancing Egocentric Narrations}
While prior egocentric VQA benchmarks~\citep{guan2024hallusionbench, li2023evaluating, wang2024videohallucer, chen2023valor} provide detailed narrations that capture a wide range of human interactions with visual elements, referred to as active objects, they often omit information about background elements, or non-active objects, that appear in the scene but are not directly interacted with. Additionally, these narrations typically lack descriptions of environmental sounds essential for comprehensive egocentric video understanding.

\looseness=-1 To address these limitations, we propose an automated pipeline to enrich egocentric narrations with visual and auditory information. As illustrated in Fig.~\ref{fig:comp_narr}, given a video $V$ with narration captions $C = \{c_1, \dots, c_n\}$ for $n$ chronologically ordered clips, along with a global video description $D$, our method first identifies active objects, denoted by $O_I = \{o_1, \dots, o_M\}$, based on objects the human interacts with in the narration captions. To detect non-active objects, we use GPT-4o~\cite{achiam2023gpt} to identify all visible objects $O_V = \{o_1, \dots, o_P\}$ from key frames sampled from each clip. The set of non-active objects is then computed as the difference $O_S \leftarrow O_V - O_I$. In parallel, to capture environmental sounds, we use Qwen2Audio~\citep{chu2024qwen2} to detect relevant audio cues from the soundtrack of each video clip, which results in an enriched set of egocentric narrations $C' = \{c'_1, \dots, c'_n\}$, where each narration $c'_i$ includes not only human actions and active objects, but also associated environmental sounds and non-active objects. Finally, a manual filtering step is applied to correct potential errors and ensure the accuracy of background object and sound descriptions.



\subsection{Generating Q/A}
\noindent{\textbf{Task Curation.}} Leveraging insights from egocentric video corpora and our enriched narrations, we curated six egocentric video-language tasks, refined from an initial pool of 20, that target core capabilities essential for egocentric understanding, including episodic reasoning, temporal inference, and human-object interaction. Each task in \method is designed to assess hallucinations in either \textit{perception} or \textit{reasoning}, with 4,000 questions allocated to each. Perception evaluates a model's ability to interpret multi-sensory inputs by recognising human actions, sounds, and visual objects in egocentric videos. In contrast, reasoning measures the model's capacity to process this information to infer knowledge, explain causality, or make decisions~\cite{fei2024video}. The selected tasks include Episodic Information Reasoning (EIR), Temporal Reasoning (TR), Human-Object Interaction (HOI), Visual Object Identification (VOI), Object State Change Detection (OSCD), and Audio Event Recognition (AER) (Additional task details are provided in Appendix~\ref{sec:tasks}). To ensure annotation consistency and quality, we developed comprehensive, task-specific guidelines outlining objectives, expected answer formats, edge cases, and annotated examples (Additional details on the annotation guidelines are provided in Appendix~\ref{sec:annotator_details}).

\noindent{\textbf{Expert Annotation.}} We employ expert annotators to generate question-answer pairs for each task (see Appendix~\ref{sec:annotator_details} for annotator details). Annotators were provided with an annotation tool, including egocentric videos, our enriched narrations, and detailed task-specific guidelines. To create hallucinated queries, annotators were instructed to apply various hallucination-inducing strategies, such as \textit{prompt injection, adversarial sampling, and temporal manipulation}. Detailed descriptions of these strategies are provided below (refer to Fig~\ref{fig:main_diag} for examples on each strategy).

\looseness=-1\noindent\textit{i) Prompt injection} is a simple yet effective technique for inducing hallucinations by exploiting a model's susceptibility to misleading or adversarial instructions~\citep{liu2024promptinjectionattackllmintegrated}. For example, given an episodic reasoning (EIR) question like ``\textit{Where did the person leave their keys?}'', we inject false information by altering the question type and replacing the referenced object with one not present in the video, producing a hallucinated version such as ``\textit{Why did the person leave their hat?}'' Extensive experiments reveal that MLLMs consistently fail to resist such attacks, lacking the ability to implicitly verify object presence before generating factually accurate responses.

\looseness=-1\noindent\textit{ii) Adversarial sampling} is employed in our benchmark to generate hallucinated queries across diverse multimodal information in egocentric videos, including human actions, sounds, and visual objects. For tasks like Hand-Object Interaction (HOI), we create hallucinated counterparts by replacing the active object (\ie the one being interacted with) with a non-active object in the scene. Using this strategy, we ensure that the hallucinated action-object pairs are scene-aware, making them harder to defend against.

\noindent\textit{iii) Manipulating temporal order} is used in our benchmark to generate hallucinated queries by altering the sequence of events defined by human-object interactions in egocentric videos. By reordering these interactions, we create mismatches between actions and the corresponding sounds they produce. This results in temporally inconsistent yet scene-plausible queries, increasing the difficulty for models in detecting hallucinations.

\noindent{\textbf{Quality Assessment.}}
To ensure the quality and consistency of the annotations, we conducted a structured quality assessment protocol involving iterative feedback and reliability checks. After initial annotation, all question-answer (QA) pairs were reviewed through a back-and-forth process between expert annotators and authors. Annotators were encouraged to flag ambiguous cases or annotation uncertainties, which were then discussed in weekly review meetings. To quantitatively assess annotation reliability, we randomly selected 1,000 QA pairs across all six tasks and had them cross-verified by expert reviewers. We measured inter-annotator agreement using Krippendorff’s Alpha, a standard metric for multi-rater agreement in benchmark construction~\citep{thrush2022winoground, li2023vlcoco}, and observed an average alpha score of \textit{0.78}, indicating substantial agreement across perception and reasoning tasks. 

\begin{table*}[t!]
\centering
\resizebox{\linewidth}{!}{
\begin{tabular}{l|c|c|c|c|ccc|ccc|c}
\toprule \toprule
\multirow{2}{*}{\textbf{Models}} & \multirow{2}{*}{\textbf{Size}} & \multirow{2}{*}{\textbf{Ego}}              & \multicolumn{2}{c|}{\textbf{Modality}}                  & \multicolumn{3}{c|}{\textbf{Reasoning Skills}}                             & \multicolumn{3}{c|}{\textbf{Perception Skills}}                                & \multirow{2}{*}{\textbf{Avg ($\uparrow$)}} \\ \cmidrule{4-11}
                &               &                           & \multicolumn{1}{c|}{Vision} & \multicolumn{1}{c|}{Audio} & \multicolumn{1}{c}{EIR ($\uparrow$)} & \multicolumn{1}{c}{TR ($\uparrow$)} & \multicolumn{1}{c|}{HOI ($\uparrow$)} & \multicolumn{1}{c}{VOI ($\uparrow$)} & \multicolumn{1}{c}{OSCD ($\uparrow$)} & \multicolumn{1}{c|}{AER ($\uparrow$)}  & \multicolumn{1}{l}{}             \\
\midrule
\multicolumn{12}{c}{\textit{Human Evaluation}} \\
\midrule
Human       &     &   & &  & 80.1\textsubscript{$\pm$0.2}  & 86.5\textsubscript{$\pm$0.2}  & 84.2\textsubscript{$\pm$0.4}  & 88.4\textsubscript{$\pm$0.5}  & 91.1\textsubscript{$\pm$0.3}  &  86.3\textsubscript{$\pm$0.2}                     & 86.1\textsubscript{$\pm$0.3}  \\

\midrule                
\multicolumn{12}{c}{\textit{Open-Source Models}}                                                                                                                                                                                                                                                           \\
\midrule
Qwen2.5VL~\citep{bai2025qwen25vltechnicalreport}       & 3B    & \textcolor{red}{$\times$}  & \textcolor{green}{\checkmark}  & \textcolor{red}{$\times$} & 50.1\textsubscript{$\pm$0.3}  & \underline{67.3}\textsubscript{$\pm$0.2}  & 54.6\textsubscript{$\pm$0.4}  & 56.3\textsubscript{$\pm$0.1}  & 51.1\textsubscript{$\pm$0.3}  & -                      & 55.8\textsubscript{$\pm$0.2}  \\
VideoLlama3~\citep{zhang2025videollama}     & 8B            & \textcolor{green}{\checkmark} & \textcolor{green}{\checkmark}  & \textcolor{red}{$\times$} & 52.1\textsubscript{$\pm$0.4}  & 59.9\textsubscript{$\pm$0.3}  & 62.7\textsubscript{$\pm$0.2}  & 63.9\textsubscript{$\pm$0.5}  & 53.2\textsubscript{$\pm$0.1}  & -                      & 58.3\textsubscript{$\pm$0.3}  \\
InternVideo~\citep{wang2025internvideo25empoweringvideomllms}     & 8B            & \textcolor{green}{\checkmark} & \textcolor{green}{\checkmark}  & \textcolor{red}{$\times$} & 51.4\textsubscript{$\pm$0.4}  & 64.3\textsubscript{$\pm$0.1}  & \underline{65.5}\textsubscript{$\pm$0.2}  & 60.8\textsubscript{$\pm$0.3}  & 51.7\textsubscript{$\pm$0.2}  & -                      & 58.7\textsubscript{$\pm$0.3}  \\
LLaVa-NEXT~\citep{li2024llavanextinterleavetacklingmultiimagevideo}      & 7B            & \textcolor{red}{$\times$} & \textcolor{green}{\checkmark}  & \textcolor{red}{$\times$} & 50.1\textsubscript{$\pm$0.2}  & 58.4\textsubscript{$\pm$0.5}  & 64.1\textsubscript{$\pm$0.1}  & 56.8\textsubscript{$\pm$0.3}  & \textbf{61.9\textsubscript{$\pm$0.4}}  & -                      & 58.2\textsubscript{$\pm$0.2}  \\
LLaVa-OV 0.5B~\citep{li2024llavaonevisioneasyvisualtask}   & 0.5B          & \textcolor{green}{\checkmark} & \textcolor{green}{\checkmark}  & \textcolor{red}{$\times$} & 51.2\textsubscript{$\pm$0.3}  & 64.5\textsubscript{$\pm$0.1}  & 61.8\textsubscript{$\pm$0.4}  & 60.5\textsubscript{$\pm$0.2}  & 52.4\textsubscript{$\pm$0.5}  & -                      & 58.1\textsubscript{$\pm$0.3}  \\
LLaVa-OV~\citep{li2024llavaonevisioneasyvisualtask}        & 7B            & \textcolor{green}{\checkmark} & \textcolor{green}{\checkmark}  & \textcolor{red}{$\times$} & 51.2\textsubscript{$\pm$0.4}  & \textbf{67.5\textsubscript{$\pm$0.2}}  & 62.9\textsubscript{$\pm$0.3}  & 58.5\textsubscript{$\pm$0.1}  & 50.3\textsubscript{$\pm$0.5}  & -                      & 58.1\textsubscript{$\pm$0.2}  \\
ImageBind-LLM~\citep{han2023imagebindllmmultimodalityinstructiontuning}   &  7B             & \textcolor{red}{$\times$} & \textcolor{green}{\checkmark}  & \textcolor{green}{\checkmark} & 55.2\textsubscript{$\pm$0.3}  & 65.6\textsubscript{$\pm$0.4}  & 61.6\textsubscript{$\pm$0.2}  & 52.9\textsubscript{$\pm$0.1}  & 51.6\textsubscript{$\pm$0.3}  & 52.2\textsubscript{$\pm$0.5}  & 57.3\textsubscript{$\pm$0.2}  \\

MiniCPM~\citep{MiniCPM2024}         & 8B            & \textcolor{red}{$\times$} & \textcolor{green}{\checkmark}  & \textcolor{green}{\checkmark} & \textbf{57.3\textsubscript{$\pm$0.4}}  & 47.3\textsubscript{$\pm$0.1}  & \textbf{66.9\textsubscript{$\pm$0.5}}  & 69.5\textsubscript{$\pm$0.3}  & \underline{58.4}\textsubscript{$\pm$0.2}  & 50.1\textsubscript{$\pm$0.4}  & \underline{58.9}\textsubscript{$\pm$0.3}  \\
VideoLlama2~\citep{cheng2024videollama}     & 7B           & \textcolor{green}{\checkmark} & \textcolor{green}{\checkmark}  & \textcolor{green}{\checkmark} & \underline{56.1}\textsubscript{$\pm$0.3}  & 38.9\textsubscript{$\pm$0.2}  & 40.2\textsubscript{$\pm$0.5}  & 41.2\textsubscript{$\pm$0.4}  & 56.8\textsubscript{$\pm$0.1}  & \textbf{52.6\textsubscript{$\pm$0.3}}  & 47.6\textsubscript{$\pm$0.2}  \\
\midrule
\multicolumn{12}{c}{\textit{Closed-Source Models}}                                                                                                                                                                                                                                                         \\
\midrule
Gemini-Pro~\citep{team2024gemini}          &  -            & - & \textcolor{green}{\checkmark}  & \textcolor{green}{\checkmark} & 51.4\textsubscript{$\pm$0.2}  & 60.8\textsubscript{$\pm$0.3}  & 61.8\textsubscript{$\pm$0.5}  & \underline{68.1}\textsubscript{$\pm$0.4}  & 56.5\textsubscript{$\pm$0.1}  & \underline{52.5}\textsubscript{$\pm$0.3}  & \textbf{59.4\textsubscript{$\pm$0.2}}  \\
GPT-4o~\citep{achiam2023gpt}          &  -            & - & \textcolor{green}{\checkmark}  & \textcolor{red}{$\times$} & 53.2\textsubscript{$\pm$0.3}  & 47.5\textsubscript{$\pm$0.2}  & 66.7\textsubscript{$\pm$0.4}  & \textbf{{73.9}\textsubscript{$\pm$0.5}}  & 58.4\textsubscript{$\pm$0.1}  & -  & 58.8\textsubscript{$\pm$0.3}  \\

\bottomrule
\end{tabular}}
\caption{\small Performance comparison of various MLLMs on \method across egocentric video-language tasks: Episodic Information Reasoning (\textbf{EIR}), Temporal Reasoning (\textbf{TR}), Human-Object Interaction (\textbf{HOI}), Visual Object Identification (\textbf{VOI}), Object State Change Detection (\textbf{OSCD}), and Audio Event Recognition (\textbf{AER}). We indicate whether the models were trained on egocentric video data and whether they leverage both vision and audio modalities. The best-performing models for each task are highlighted in \textbf{bold}, while the second-best scores are \underline{underlined}.}
\label{tab:main_result}
\end{table*}

\section{Experimental Setup}

We first describe the baselines used to evaluate hallucination performance and then outline the human evaluation setup. 

\noindent{\textbf{Baselines.}} We benchmark a range of MLLMs, including eight open-weight and closed-source models, such as Gemini-1.5~\citep{team2024gemini} and GPT-4o~\citep{achiam2023gpt}. These models are selected to cover a wide variety of factors, including \textit{model size} (LLaVa-OV~\citep{li2024llavaonevisioneasyvisualtask} contains 0.5B parameters, whereas VideoLLaMA2~\citep{cheng2024videollama} consists of 7B parameters). They also vary in the \textit{video type} used during training (ImageBind-LLM~\citep{han2023imagebindllmmultimodalityinstructiontuning} is trained solely on exocentric videos, while VideoLLaMA3~\citep{zhang2025videollama} and InternVideo~\citep{wang2025internvideo25empoweringvideomllms} are jointly trained on both exocentric and egocentric videos). Finally, the models differ in their \textit{multisensory input capabilities} — LLaVa-Next~\citep{li2024llavanextinterleavetacklingmultiimagevideo} and LLaVa-OV~\citep{li2024llavaonevisioneasyvisualtask} process videos without audio, in contrast to models like Gemini-1.5~\citep{team2024gemini}, which process both video and audio signals (see Appendix~\ref{sec:baselines} for additional details).

\looseness=-1\noindent\textbf{Hallucination Evaluation.} We conduct separate evaluations for both close-ended and open-ended questions. For close-ended questions, which require binary yes/no answers, we follow prior video hallucination benchmarks such as VideoHallucer~\citep{wang2024videohallucer} by applying string matching to convert model responses into either ``\textit{Yes}'' or ``\textit{No}.'' For open-ended questions, we adopt a two-step approach: first, we determine whether the model implicitly assumes the presence of an object using an LLM-as-judge framework~\citep{zheng2023judging} with GPT-4o~\citep{achiam2023gpt} (to reduce model bias, we also use Gemini-Pro for LLM-as-judge); second, we independently assess the factual correctness of the response. Consistent with previous hallucination benchmarks, we report accuracy as the primary metric, where lower accuracy indicates a higher degree of hallucinations. 

\noindent{\textbf{Human Evaluation.}} We recruited three English-proficient individuals to evaluate our benchmark, where each individual had strong foundational knowledge of computer vision. To reduce potential evaluator bias, we randomized the order of the question-answer pairs, ensuring that correct and hallucinated responses did not appear consecutively. Inter-annotator reliability was measured using the Pearson correlation coefficient, yielding a moderate agreement score of \textit{0.58}.
\section{Results}
\subsection{Main Results}
\label{res:main_result}
We benchmark ten state-of-the-art MLLMs on \method and present the results in Table~\ref{tab:main_result}. Below, we summarize the key findings:

\looseness=-1\noindent{\textit{i) \method presents a significant challenge}}, exposing the vulnerability of current MLLMs to hallucination. We find that existing models struggle to defend against hallucinations induced by \method. For instance, the best-performing model, Gemini-Pro, achieves 59.4\% accuracy, while human performance on the benchmark is 86.1\%, revealing a gap of \textit{26.7\%}.

\noindent{\textit{ii) Minimal performance gap between open- and closed-weight model.}} Unlike other benchmarks, \method reveals only a small performance gap between open- and closed-weight models. In Table~\ref{tab:main_result}, we show that the best open-weight model, VideoLlama3, achieves an accuracy of 58.3\%, while the best closed-weight model, Gemini-Pro, reaches 59.4\%, a marginal difference of \textit{1\%}.

\looseness=-1\noindent{\textit{iii) Minimal performance gap between small and large MLLMs.}} Unlike conventional benchmarks where larger models typically outperform smaller ones, \method reveals that model size alone does not consistently mitigate hallucinations, \eg the small LLaVA-OV 0.5B model achieves 58.1\% average accuracy, matching the performance of its larger counterpart, LLaVA-OV 7B, suggesting that the hallucinations introduced by \method are not easily mitigated by scaling model size.

\looseness=-1\noindent{\textit{iv) MLLMs hallucinate less on perception-based tasks than on reasoning tasks.}} As shown in Table~\ref{tab:main_result}, MLLMs hallucinate less on perception-based tasks (Visual Object Identification (VOI) and Audio Event Recognition (AER)) compared to reasoning tasks (Temporal Reasoning (TR) and Episodic Information Reasoning (EIR)). For example, the best-performing model, Gemini-Pro, achieves 68.1\% accuracy on VOI and 58.3\% on AER, but only 60.8\% on TR and 51.4\% on EIR—\textit{a gap of over 7\%}. This suggests that hallucinations are more prevalent when models are asked to perform complex reasoning rather than perception.


\begin{table}[]
\small
\setlength{\tabcolsep}{8.0pt}
\centering
\begin{tabular}{lccc}
\toprule \toprule
\textbf{Models} & PI  ($\uparrow$) & AS  ($\uparrow$) & MTO ($\uparrow$) \\ \midrule
\multicolumn{4}{c}{\textit{Open-Weight Models}}  \\ \midrule
ImageBind-LLM &  54.5\textsubscript{$\pm$0.3}   &  61.6\textsubscript{$\pm$0.4}  &  65.6\textsubscript{$\pm$0.2}   \\
Qwen2.5VL & 53.2\textsubscript{$\pm$0.2}   & 52.8\textsubscript{$\pm$0.3}   &   67.3\textsubscript{$\pm$0.5}  \\
VideoLlama3 &  60.1\textsubscript{$\pm$0.4}  &  66.0\textsubscript{$\pm$0.2}  &  59.9\textsubscript{$\pm$0.3}   \\
LLaVa-NEXT &  58.0\textsubscript{$\pm$0.1}  &  65.3\textsubscript{$\pm$0.5}  &  58.4\textsubscript{$\pm$0.3}   \\
LLaVa-OV 0.5B & 56.5\textsubscript{$\pm$0.3}   &  57.2\textsubscript{$\pm$0.4}  &   64.5\textsubscript{$\pm$0.2}  \\
LLaVa-OV &  54.8\textsubscript{$\pm$0.2}  & 56.8\textsubscript{$\pm$0.3}   &  67.5\textsubscript{$\pm$0.4}   \\
MiniCPMo-2.6 &   58.4\textsubscript{$\pm$0.5} & 51.0\textsubscript{$\pm$0.2}   &   47.3\textsubscript{$\pm$0.3}  \\
VideoLlama2 &  58.9\textsubscript{$\pm$0.3}  &  51.0\textsubscript{$\pm$0.4} & 38.9\textsubscript{$\pm$0.2}    \\ \midrule
\multicolumn{4}{c}{\textit{Closed-Source Models}} \\ \midrule
Gemini-Pro &  53.9\textsubscript{$\pm$0.4}  & 64.9\textsubscript{$\pm$0.2}   &  60.8\textsubscript{$\pm$0.5}   \\ 
GPT-4o &  54.2\textsubscript{$\pm$0.3}  & 62.1\textsubscript{$\pm$0.1}   &  59.7\textsubscript{$\pm$0.3}   \\\bottomrule
\end{tabular}
\caption{\small Performance comparison of various MLLMs across diverse hallucination-inducing strategies employed in \method, including prompt injection (PI), Adversarial Sampling (AS), and Manipulating Temporal Order (MTO).}
\label{tab:hallu_table}
\end{table}

\begin{figure}
    \centering
    \includegraphics[width=1.0\columnwidth]{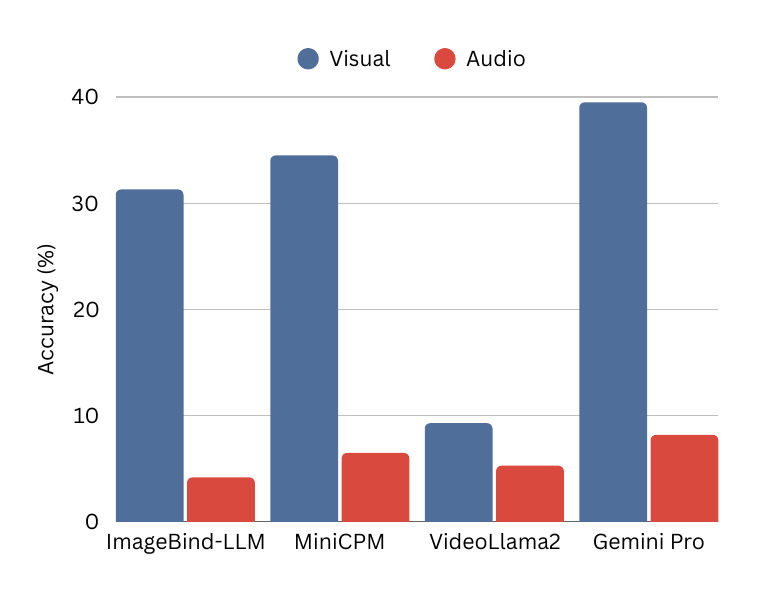}
    \caption{\small Performance comparison on confounding pairs generated from the videos Q/A sourced from \method across visual and audio modality.}
    \label{fig:gauss_video}
\end{figure}

\subsection{Ablation On Hallucination Inducing Strategies}

\label{res:his}
Building on these findings, we further examine how different hallucination-inducing strategies affect MLLM performance on \method. Table~\ref{tab:hallu_table} compares the performance of various MLLMs under different hallucination-inducing strategies employed in the \method. Overall, models tend to perform close to random guess across all strategies, highlighting their consistent vulnerability to hallucinations in egocentric video understanding. Among open-weight models, MiniCPM and VideoLlama2 perform the worst, particularly under the Manipulating Temporal Order (MTO) strategy, where their scores drop to 47.3\% and 38.9\%, respectively, indicating significant difficulty in understanding chronological ordering in unique egocentric events. For closed-weight models, Gemini-Pro and GPT-4o perform reasonably well compared to open-weight models but remain susceptible to hallucinations induced by Prompt Injection (PI), where they achieve the lowest score (53.9\%), indicating that these MLLMs are vulnerable to misleading prompts, likely due to learned biases from pretraining data that make them more susceptible to hallucinated inputs.



\subsection{Which modality does MLLMs attend to?}
Motivated by the near-random performance of current MLLMs on our benchmark, we further investigate which modality (audio or visual) these models primarily attend to while understanding egocentric videos. We conduct an experiment by randomly selecting 200 video clips from \method and generating confounding pairs to isolate the contribution of each modality. For the audio modality, we synthetically add unrelated background sounds; for the visual modality, we replace the main object in the query with a random object. A model’s response is considered correct only if it answers both versions of the confounding pair correctly. As shown in Fig.~\ref{fig:gauss_video}, when evaluated on MLLMs that process both modalities, we find a significant drop in performance below 50\%, on both types of perturbations. Notably, the \textit{performance degradation is more severe for audio}, with a 32\% drop for Gemini and 28\% for MiniCPM. These results demonstrate that while MLLMs struggle with both modalities, they especially fail to leverage audio cues, instead relying heavily on language priors, leading to hallucinated responses.


\begin{figure}
    \centering
    \includegraphics[width=1.0\linewidth]{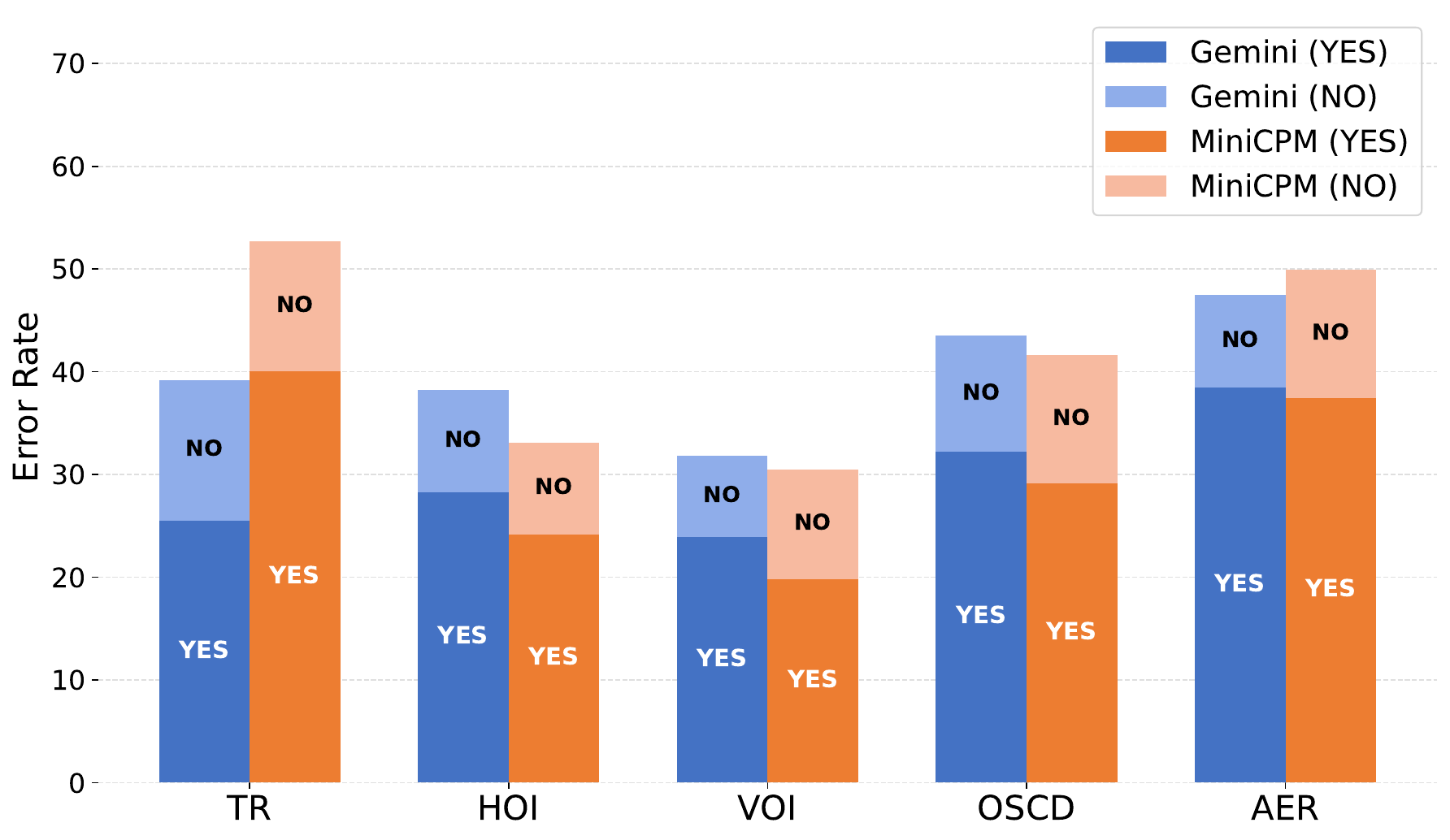}
    \vspace{-0.05in}
    \caption{\small Distribution of ``\textit{Yes}'' and ``\textit{No}'' responses of Gemini-1.5 Pro and MiniCPM in the hallucinated responses for closed-ended questions. We observe that the model is inclined towards affirmative responses in hallucinated outputs.}
    \vspace{-0.05in}
    \label{fig:quan_y_n}
\end{figure}

\subsection{Error Analysis}
\label{res:er}
Next, we conduct a detailed error analysis with a focus on response biases and the failure cases.

\looseness=-1{\noindent \textbf{Yes/No Bias.}} Fig.~\ref{fig:quan_y_n} presents a quantitative analysis of how often Gemini-1.5 Pro and MiniCPM respond with ``\textit{Yes}'' or ``\textit{No}'' when generating hallucinated responses in closed-ended tasks within \method. We observe that despite differing hallucination rates, both models exhibit a significantly higher proportion of ``\textit{Yes}'' responses compared to ``\textit{No}'' across various tasks, \eg in egocentric video-language tasks such as Visual-Object Identification (VOI), where both models show similar hallucination rates, we find that they still demonstrate a strong bias toward ``\textit{Yes}'' responses. A similar pattern emerges in Temporal Reasoning (TR), where the models differ in their hallucination rates but still predominantly produce ``\textit{Yes}'' responses. This trend remains consistent across other tasks, as shown in Fig.~\ref{fig:quan_y_n}, indicating the models' inclination toward affirmative responses in hallucinated outputs.

\begin{figure}[t]
    \centering
    \includegraphics[width=0.9\linewidth]{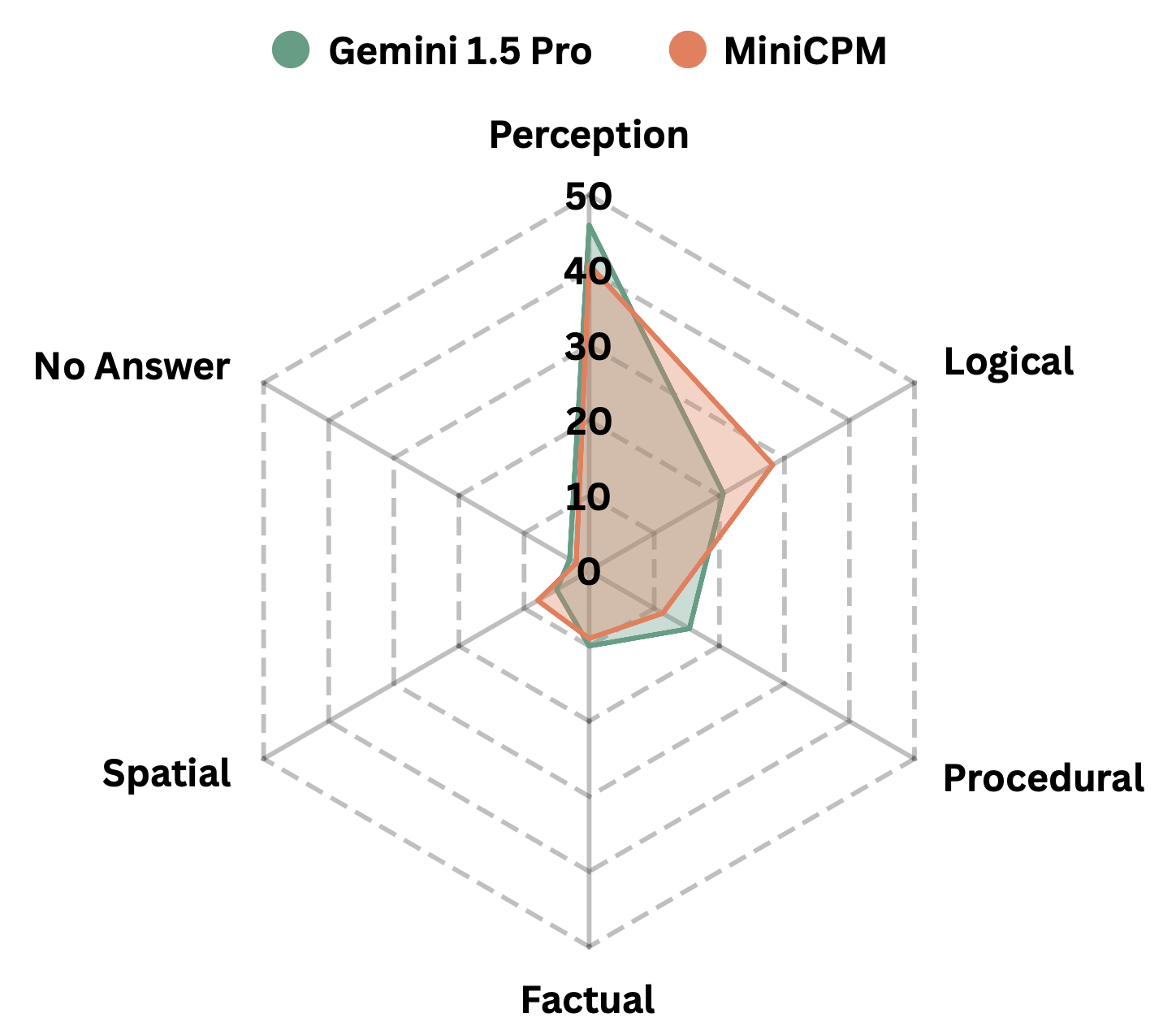}
    \vspace{-0.05in}
    \caption{\small An illustration of the different types of errors observed in incorrect responses from Gemini 1.5 Pro~\citep{team2024gemini} and MiniCPM~\citep{MiniCPM2024}. Additional details on the various error types can be found in Appendix~\ref{sec:open_ended_error_analysis}.}
    \vspace{-0.05in}
    \label{fig:sk_pie}
\end{figure}

{\noindent \textbf{Finegrained Error Analysis.}} We conduct a manual error analysis on 1,000 incorrect responses, representing 12.5\% of the total benchmark samples, uniformly sampled across all six tasks in \method. Fig.~\ref{fig:sk_pie} presents a detailed breakdown of the different types of errors observed in responses generated by Gemini 1.5 Pro~\citep{team2024gemini} and MiniCPM~\citep{MiniCPM2024} on \method. The primary source of errors for both models is \textit{perception}, accounting for 48.6\% of Gemini 1.5 Pro's mistakes and 43.7\% of MiniCPM's. This is largely driven by hallucination-inducing questions in \method, revealing the models' difficulty in accurately perceiving entities in the video before generating factually grounded responses. In addition, \textit{logical} and \textit{procedural} errors make up a substantial share of the failures, indicating that even when models identify relevant entities correctly, they often fall short in applying the complex reasoning needed for accurate answers. Overall, this analysis underscores the critical need for improved perceptual understanding in egocentric video tasks.


\section{Conclusion}

In this paper, we introduced \method, the first comprehensive benchmark specifically designed to evaluate hallucination in MLLMs within egocentric video understanding. Our benchmark features over 1,400 egocentric videos and 8,000 carefully annotated question-answer pairs designed to systematically trigger and assess hallucinations across diverse scenarios involving audio and visual perception and complex reasoning. Experimental results across ten SOTA MLLMs reveal significant vulnerabilities, demonstrating that current models, regardless of scale or training modality, are highly susceptible to hallucinations, achieving accuracies close to random guessing. By introducing novel hallucination inducing techniques, \method provides insights into the MLLM's limitations and offers a roadmap for future research. 


\newpage
\section{Limitation and Future Work}
In this section, we highlight a few limitations and future directions:
   \begin{itemize}
    \item Our benchmark, \method reveals that existing Multimodal Large Language Models (MLLMs) exhibit a high rate of hallucination when evaluated on egocentric video understanding tasks. In future work, we plan to develop robust hallucination mitigation strategies tailored specifically for this domain.

    \item While the current version of our benchmark evaluates model performance on visual and non-speech auditory cues (e.g., background sounds) in egocentric videos, it does not yet cover speech signals. As egocentric videos often contain conversations, we aim to extend our benchmark to include the speech modality in future iterations, enabling more comprehensive evaluations and analysis.
\end{itemize}

\newpage
\bibliography{acl_latex}

\begin{thebibliography}{51}
\providecommand{\natexlab}[1]{#1}

\bibitem[{Achiam et~al.(2023)Achiam, Adler, Agarwal, Ahmad, Akkaya, Aleman, Almeida, Altenschmidt, Altman, Anadkat et~al.}]{achiam2023gpt}
Josh Achiam, Steven Adler, Sandhini Agarwal, Lama Ahmad, Ilge Akkaya, Florencia~Leoni Aleman, Diogo Almeida, Janko Altenschmidt, Sam Altman, Shyamal Anadkat, et~al. 2023.
\newblock Gpt-4 technical report.
\newblock \emph{arXiv}.

\bibitem[{Bai et~al.(2025)Bai, Chen, Liu, Wang, Ge, Song, Dang, Wang, Wang, Tang, Zhong, Zhu, Yang, Li, Wan, Wang, Ding, Fu, Xu, Ye, Zhang, Xie, Cheng, Zhang, Yang, Xu, and Lin}]{bai2025qwen25vltechnicalreport}
Shuai Bai, Keqin Chen, Xuejing Liu, Jialin Wang, Wenbin Ge, Sibo Song, Kai Dang, Peng Wang, Shijie Wang, Jun Tang, Humen Zhong, Yuanzhi Zhu, Mingkun Yang, Zhaohai Li, Jianqiang Wan, Pengfei Wang, Wei Ding, Zheren Fu, Yiheng Xu, Jiabo Ye, Xi~Zhang, Tianbao Xie, Zesen Cheng, Hang Zhang, Zhibo Yang, Haiyang Xu, and Junyang Lin. 2025.
\newblock Qwen2.5-vl technical report.

\bibitem[{B{\"a}rmann and Waibel(2022)}]{barmann2022did}
Leonard B{\"a}rmann and Alex Waibel. 2022.
\newblock Where did i leave my keys?-episodic-memory-based question answering on egocentric videos.
\newblock In \emph{CVPR}.

\bibitem[{Chen et~al.(2024{\natexlab{a}})Chen, Ashutosh, Girdhar, Harwath, and Grauman}]{chen2024soundingactionslearningactionssound}
Changan Chen, Kumar Ashutosh, Rohit Girdhar, David Harwath, and Kristen Grauman. 2024{\natexlab{a}}.
\newblock Soundingactions: Learning how actions sound from narrated egocentric videos.

\bibitem[{Chen et~al.(2024{\natexlab{b}})Chen, Peng, Baid, Xue, Hsu, Harwath, and Grauman}]{chen2024action2soundambientawaregenerationaction}
Changan Chen, Puyuan Peng, Ami Baid, Zihui Xue, Wei-Ning Hsu, David Harwath, and Kristen Grauman. 2024{\natexlab{b}}.
\newblock Action2sound: Ambient-aware generation of action sounds from egocentric videos.

\bibitem[{Chen et~al.(2023)Chen, He, Guo, Zhu, Wang, Tang, and Liu}]{chen2023valor}
Sihan Chen, Xingjian He, Longteng Guo, Xinxin Zhu, Weining Wang, Jinhui Tang, and Jing Liu. 2023.
\newblock Valor: Vision-audio-language omni-perception pretraining model and dataset.
\newblock \emph{arXiv}.

\bibitem[{Chen et~al.(2024{\natexlab{c}})Chen, Wang, Xue, Zhang, Yang, Li, Shen, Liang, Gu, and Chen}]{chen2024unified}
Xiang Chen, Chenxi Wang, Yida Xue, Ningyu Zhang, Xiaoyan Yang, Qiang Li, Yue Shen, Lei Liang, Jinjie Gu, and Huajun Chen. 2024{\natexlab{c}}.
\newblock Unified hallucination detection for multimodal large language models.
\newblock \emph{arXiv}.

\bibitem[{Cheng et~al.(2024)Cheng, Leng, Zhang, Xin, Li, Chen, Zhu, Zhang, Luo, Zhao et~al.}]{cheng2024videollama}
Zesen Cheng, Sicong Leng, Hang Zhang, Yifei Xin, Xin Li, Guanzheng Chen, Yongxin Zhu, Wenqi Zhang, Ziyang Luo, Deli Zhao, et~al. 2024.
\newblock Videollama 2: Advancing spatial-temporal modeling and audio understanding in video-llms.
\newblock \emph{arXiv}.

\bibitem[{Chu et~al.(2024)Chu, Xu, Yang, Wei, Wei, Guo, Leng, Lv, He, Lin et~al.}]{chu2024qwen2}
Yunfei Chu, Jin Xu, Qian Yang, Haojie Wei, Xipin Wei, Zhifang Guo, Yichong Leng, Yuanjun Lv, Jinzheng He, Junyang Lin, et~al. 2024.
\newblock Qwen2-audio technical report.
\newblock \emph{arXiv}.

\bibitem[{Cui et~al.(2023)Cui, Zhou, Yang, Wu, Zhang, Zou, and Yao}]{cui2023holistic}
Chenhang Cui, Yiyang Zhou, Xinyu Yang, Shirley Wu, Linjun Zhang, James Zou, and Huaxiu Yao. 2023.
\newblock Holistic analysis of hallucination in gpt-4v (ision): Bias and interference challenges.
\newblock \emph{arXiv}.

\bibitem[{Damen et~al.(2022)Damen, Doughty, Farinella, Furnari, Ma, Kazakos, Moltisanti, Munro, Perrett, Price, and Wray}]{Damen2022RESCALING}
Dima Damen, Hazel Doughty, Giovanni~Maria Farinella, Antonino Furnari, Jian Ma, Evangelos Kazakos, Davide Moltisanti, Jonathan Munro, Toby Perrett, Will Price, and Michael Wray. 2022.
\newblock \href {https://doi.org/10.1007/s11263-021-01531-2} {Rescaling egocentric vision: Collection, pipeline and challenges for epic-kitchens-100}.
\newblock \emph{International Journal of Computer Vision (IJCV)}, 130:33–55.

\bibitem[{Di and Xie(2024)}]{di2024grounded}
Shangzhe Di and Weidi Xie. 2024.
\newblock Grounded question-answering in long egocentric videos.
\newblock In \emph{CVPR}.

\bibitem[{Dubey et~al.(2024)Dubey, Jauhri, Pandey, Kadian, Al-Dahle, Letman, Mathur, Schelten, Yang, Fan et~al.}]{dubey2024llama}
Abhimanyu Dubey, Abhinav Jauhri, Abhinav Pandey, Abhishek Kadian, Ahmad Al-Dahle, Aiesha Letman, Akhil Mathur, Alan Schelten, Amy Yang, Angela Fan, et~al. 2024.
\newblock The llama 3 herd of models.
\newblock \emph{arXiv}.

\bibitem[{Dunnhofer et~al.(2022)Dunnhofer, Furnari, Farinella, and Micheloni}]{TREK150ijcv}
Matteo Dunnhofer, Antonino Furnari, Giovanni~Maria Farinella, and Christian Micheloni. 2022.
\newblock Visual object tracking in first person vision.
\newblock \emph{IJCV}.

\bibitem[{Fei et~al.(2024)Fei, Wu, Ji, Zhang, Zhang, Lee, and Hsu}]{fei2024video}
Hao Fei, Shengqiong Wu, Wei Ji, Hanwang Zhang, Meishan Zhang, Mong-Li Lee, and Wynne Hsu. 2024.
\newblock Video-of-thought: Step-by-step video reasoning from perception to cognition.
\newblock \emph{arXiv}.

\bibitem[{Fu et~al.(2024)Fu, Dai, Luo, Li, Ren, Zhang, Wang, Zhou, Shen, Zhang et~al.}]{fu2024video}
Chaoyou Fu, Yuhan Dai, Yongdong Luo, Lei Li, Shuhuai Ren, Renrui Zhang, Zihan Wang, Chenyu Zhou, Yunhang Shen, Mengdan Zhang, et~al. 2024.
\newblock Video-mme: The first-ever comprehensive evaluation benchmark of multi-modal llms in video analysis.
\newblock \emph{arXiv}.

\bibitem[{Grauman et~al.(2022)Grauman, Westbury, Byrne, Chavis, Furnari, Girdhar, Hamburger, Jiang, Liu, Liu et~al.}]{grauman2022ego4d}
Kristen Grauman, Andrew Westbury, Eugene Byrne, Zachary Chavis, Antonino Furnari, Rohit Girdhar, Jackson Hamburger, Hao Jiang, Miao Liu, Xingyu Liu, et~al. 2022.
\newblock Ego4d: Around the world in 3,000 hours of egocentric video.
\newblock In \emph{CVPR}.

\bibitem[{Grauman et~al.(2024)Grauman, Westbury, Torresani, Kitani, Malik, Afouras, Ashutosh, Baiyya, Bansal, Boote et~al.}]{grauman2024ego}
Kristen Grauman, Andrew Westbury, Lorenzo Torresani, Kris Kitani, Jitendra Malik, Triantafyllos Afouras, Kumar Ashutosh, Vijay Baiyya, Siddhant Bansal, Bikram Boote, et~al. 2024.
\newblock Ego-exo4d: Understanding skilled human activity from first-and third-person perspectives.
\newblock In \emph{CVPR}.

\bibitem[{Guan et~al.(2024)Guan, Liu, Wu, Xian, Li, Liu, Wang, Chen, Huang, Yacoob et~al.}]{guan2024hallusionbench}
Tianrui Guan, Fuxiao Liu, Xiyang Wu, Ruiqi Xian, Zongxia Li, Xiaoyu Liu, Xijun Wang, Lichang Chen, Furong Huang, Yaser Yacoob, et~al. 2024.
\newblock Hallusionbench: an advanced diagnostic suite for entangled language hallucination and visual illusion in large vision-language models.
\newblock In \emph{CVPR}.

\bibitem[{Han et~al.(2023)Han, Zhang, Shao, Gao, Xu, Xiao, Zhang, Liu, Wen, Guo, Lu, Ren, Wen, Chen, Yue, Li, and Qiao}]{han2023imagebindllmmultimodalityinstructiontuning}
Jiaming Han, Renrui Zhang, Wenqi Shao, Peng Gao, Peng Xu, Han Xiao, Kaipeng Zhang, Chris Liu, Song Wen, Ziyu Guo, Xudong Lu, Shuai Ren, Yafei Wen, Xiaoxin Chen, Xiangyu Yue, Hongsheng Li, and Yu~Qiao. 2023.
\newblock Imagebind-llm: Multi-modality instruction tuning.

\bibitem[{Hatano et~al.(2024)Hatano, Hachiuma, Fujii, and Saito}]{hatano2024multimodalcrossdomainfewshotlearning}
Masashi Hatano, Ryo Hachiuma, Ryo Fujii, and Hideo Saito. 2024.
\newblock Multimodal cross-domain few-shot learning for egocentric action recognition.

\bibitem[{Huang et~al.(2024)Huang, Liu, Guo, and Gong}]{huang2024visual}
Wen Huang, Hongbin Liu, Minxin Guo, and Neil~Zhenqiang Gong. 2024.
\newblock Visual hallucinations of multi-modal large language models.
\newblock \emph{arXiv}.

\bibitem[{Islam et~al.(2024)Islam, Ho, Yang, Nagarajan, Torresani, and Bertasius}]{islam2024video}
Md~Mohaiminul Islam, Ngan Ho, Xitong Yang, Tushar Nagarajan, Lorenzo Torresani, and Gedas Bertasius. 2024.
\newblock Video recap: Recursive captioning of hour-long videos.
\newblock In \emph{CVPR}.

\bibitem[{Jia et~al.(2024)Jia, Liu, Jiang, Ananthabhotla, Rehg, Ithapu, and Gao}]{Jia_2024_CVPR}
Wenqi Jia, Miao Liu, Hao Jiang, Ishwarya Ananthabhotla, James~M. Rehg, Vamsi~Krishna Ithapu, and Ruohan Gao. 2024.
\newblock The audio-visual conversational graph: From an egocentric-exocentric perspective.
\newblock In \emph{CVPR}.

\bibitem[{Kaul et~al.(2024)Kaul, Li, Yang, Dukler, Swaminathan, Taylor, and Soatto}]{kaul2024throne}
Prannay Kaul, Zhizhong Li, Hao Yang, Yonatan Dukler, Ashwin Swaminathan, CJ~Taylor, and Stefano Soatto. 2024.
\newblock Throne: An object-based hallucination benchmark for the free-form generations of large vision-language models.
\newblock In \emph{CVPR}.

\bibitem[{Kim et~al.(2024)Kim, Huang, Xian, Hilliges, Gool, and Wang}]{kim2024palmpredictingactionslanguage}
Sanghwan Kim, Daoji Huang, Yongqin Xian, Otmar Hilliges, Luc~Van Gool, and Xi~Wang. 2024.
\newblock Palm: Predicting actions through language models.

\bibitem[{Li et~al.(2024{\natexlab{a}})Li, Zhang, Guo, Zhang, Li, Zhang, Zhang, Zhang, Li, Liu, and Li}]{li2024llavaonevisioneasyvisualtask}
Bo~Li, Yuanhan Zhang, Dong Guo, Renrui Zhang, Feng Li, Hao Zhang, Kaichen Zhang, Peiyuan Zhang, Yanwei Li, Ziwei Liu, and Chunyuan Li. 2024{\natexlab{a}}.
\newblock Llava-onevision: Easy visual task transfer.

\bibitem[{Li et~al.(2024{\natexlab{b}})Li, Zhang, Zhang, Zhang, Li, Li, Ma, and Li}]{li2024llavanextinterleavetacklingmultiimagevideo}
Feng Li, Renrui Zhang, Hao Zhang, Yuanhan Zhang, Bo~Li, Wei Li, Zejun Ma, and Chunyuan Li. 2024{\natexlab{b}}.
\newblock Llava-next-interleave: Tackling multi-image, video, and 3d in large multimodal models.

\bibitem[{Li et~al.(2024{\natexlab{c}})Li, Wang, He, Li, Wang, Liu, Wang, Xu, Chen, Luo, Wang, and Qiao}]{li2024mvbenchcomprehensivemultimodalvideo}
Kunchang Li, Yali Wang, Yinan He, Yizhuo Li, Yi~Wang, Yi~Liu, Zun Wang, Jilan Xu, Guo Chen, Ping Luo, Limin Wang, and Yu~Qiao. 2024{\natexlab{c}}.
\newblock Mvbench: A comprehensive multi-modal video understanding benchmark.

\bibitem[{Li et~al.(2023{\natexlab{a}})}]{li2023vlcoco}
Xinyu Li et~al. 2023{\natexlab{a}}.
\newblock Vl-coco: A benchmark for visio-linguistic compositional reasoning.
\newblock In \emph{ICCV}.

\bibitem[{Li et~al.(2023{\natexlab{b}})Li, Du, Zhou, Wang, Zhao, and Wen}]{li2023evaluating}
Yifan Li, Yifan Du, Kun Zhou, Jinpeng Wang, Wayne~Xin Zhao, and Ji-Rong Wen. 2023{\natexlab{b}}.
\newblock Evaluating object hallucination in large vision-language models.
\newblock \emph{arXiv}.

\bibitem[{Liu et~al.(2024)Liu, Deng, Li, Wang, Wang, Wang, Zhang, Liu, Wang, Zheng, and Liu}]{liu2024promptinjectionattackllmintegrated}
Yi~Liu, Gelei Deng, Yuekang Li, Kailong Wang, Zihao Wang, Xiaofeng Wang, Tianwei Zhang, Yepang Liu, Haoyu Wang, Yan Zheng, and Yang Liu. 2024.
\newblock Prompt injection attack against llm-integrated applications.

\bibitem[{Luo et~al.(2024)Luo, Xue, Dimakis, and Grauman}]{luo2024shoesliftingegocentricperspective}
Mi~Luo, Zihui Xue, Alex Dimakis, and Kristen Grauman. 2024.
\newblock Put myself in your shoes: Lifting the egocentric perspective from exocentric videos.

\bibitem[{Mangalam et~al.(2023)Mangalam, Akshulakov, and Malik}]{mangalam2023egoschema}
Karttikeya Mangalam, Raiymbek Akshulakov, and Jitendra Malik. 2023.
\newblock Egoschema: A diagnostic benchmark for very long-form video language understanding.
\newblock \emph{NeurIPS}.

\bibitem[{OpenBMB(2024)}]{MiniCPM2024}
OpenBMB. 2024.
\newblock \href {https://openbmb.notion.site/MiniCPM-o-2-6-A-GPT-4o-Level-MLLM-for-Vision-Speech-and-Multimodal-Live-Streaming-on-Your-Phone-185ede1b7a558042b5d5e45e6b237da9} {Minicpm-o 2.6: A gpt-4o level mllm for vision, speech, and multimodal live streaming on your phone}.
\newblock Accessed: 2025-03-07.

\bibitem[{Poleg et~al.(2016)Poleg, Ephrat, Peleg, and Arora}]{poleg_wacv16_compactcnn}
Yair Poleg, Ariel Ephrat, Shmuel Peleg, and Chetan Arora. 2016.
\newblock Compact cnn for indexing egocentric videos.
\newblock In \emph{WACV}.

\bibitem[{Shen et~al.(2024)Shen, Dudley, and Kristensson}]{shen2024encode}
Junxiao Shen, John~J Dudley, and Per~Ola Kristensson. 2024.
\newblock Encode-store-retrieve: Augmenting human memory through language-encoded egocentric perception.
\newblock In \emph{IEEE ISMAR}.

\bibitem[{Su et~al.(2024)Su, Liu, and Shlizerman}]{su2024icml}
Kun Su, Xiulong Liu, and Eli Shlizerman. 2024.
\newblock From vision to audio and beyond: a unified model for audio-visual representation and generation.
\newblock In \emph{ICML}.

\bibitem[{Sun et~al.(2023)Sun, Shen, Cao, Liu, Li, Shen, Gan, Gui, Wang, Yang et~al.}]{sun2023aligning}
Zhiqing Sun, Sheng Shen, Shengcao Cao, Haotian Liu, Chunyuan Li, Yikang Shen, Chuang Gan, Liang-Yan Gui, Yu-Xiong Wang, Yiming Yang, et~al. 2023.
\newblock Aligning large multimodal models with factually augmented rlhf.
\newblock \emph{arXiv}.

\bibitem[{Team et~al.(2024)Team, Georgiev, Lei, Burnell, Bai, Gulati, Tanzer, Vincent, Pan, Wang et~al.}]{team2024gemini}
Gemini Team, Petko Georgiev, Ving~Ian Lei, Ryan Burnell, Libin Bai, Anmol Gulati, Garrett Tanzer, Damien Vincent, Zhufeng Pan, Shibo Wang, et~al. 2024.
\newblock Gemini 1.5: Unlocking multimodal understanding across millions of tokens of context.
\newblock \emph{arXiv}.

\bibitem[{Thrush et~al.(2022)}]{thrush2022winoground}
Tristan Thrush et~al. 2022.
\newblock Winoground: Probing vision and language models for visio-linguistic compositionality.
\newblock In \emph{CVPR}.

\bibitem[{Wang et~al.(2023)Wang, Wang, Xu, Zhang, Gu, Jia, Yan, Zhang, and Sang}]{wang2023llm}
Junyang Wang, Yuhang Wang, Guohai Xu, Jing Zhang, Yukai Gu, Haitao Jia, Ming Yan, Ji~Zhang, and Jitao Sang. 2023.
\newblock An llm-free multi-dimensional benchmark for mllms hallucination evaluation.
\newblock \emph{arXiv}.

\bibitem[{Wang et~al.(2024{\natexlab{a}})Wang, Bai, Tan, Wang, Fan, Bai, Chen, Liu, Wang, Ge et~al.}]{wang2024qwen2}
Peng Wang, Shuai Bai, Sinan Tan, Shijie Wang, Zhihao Fan, Jinze Bai, Keqin Chen, Xuejing Liu, Jialin Wang, Wenbin Ge, et~al. 2024{\natexlab{a}}.
\newblock Qwen2-vl: Enhancing vision-language model's perception of the world at any resolution.
\newblock \emph{arXiv}.

\bibitem[{Wang et~al.(2025)Wang, Li, Yan, He, Yu, Zeng, Wang, Ma, Huang, Gao, Dou, Chen, Wang, Qiao, Wang, and Wang}]{wang2025internvideo25empoweringvideomllms}
Yi~Wang, Xinhao Li, Ziang Yan, Yinan He, Jiashuo Yu, Xiangyu Zeng, Chenting Wang, Changlian Ma, Haian Huang, Jianfei Gao, Min Dou, Kai Chen, Wenhai Wang, Yu~Qiao, Yali Wang, and Limin Wang. 2025.
\newblock Internvideo2.5: Empowering video mllms with long and rich context modeling.

\bibitem[{Wang et~al.(2024{\natexlab{b}})Wang, Wang, Zhao, Xie, and Zheng}]{wang2024videohallucer}
Yuxuan Wang, Yueqian Wang, Dongyan Zhao, Cihang Xie, and Zilong Zheng. 2024{\natexlab{b}}.
\newblock Videohallucer: Evaluating intrinsic and extrinsic hallucinations in large video-language models.
\newblock \emph{arXiv}.

\bibitem[{Wu et~al.(2024)Wu, Chen, Pan, Liu, Liu, Dai, Gao, Ma, Wu, Wang, Xie, Wu, Hu, Wang, Sun, Li, Piao, Guan, Liu, Xie, You, Dong, Yu, Zhang, Zhao, Wang, and Ruan}]{wu2024deepseekvl2mixtureofexpertsvisionlanguagemodels}
Zhiyu Wu, Xiaokang Chen, Zizheng Pan, Xingchao Liu, Wen Liu, Damai Dai, Huazuo Gao, Yiyang Ma, Chengyue Wu, Bingxuan Wang, Zhenda Xie, Yu~Wu, Kai Hu, Jiawei Wang, Yaofeng Sun, Yukun Li, Yishi Piao, Kang Guan, Aixin Liu, Xin Xie, Yuxiang You, Kai Dong, Xingkai Yu, Haowei Zhang, Liang Zhao, Yisong Wang, and Chong Ruan. 2024.
\newblock Deepseek-vl2: Mixture-of-experts vision-language models for advanced multimodal understanding.

\bibitem[{Xu et~al.(2023)Xu, Lan, Xie, Chen, and Lu}]{xu2023retrieval}
Jiaqi Xu, Cuiling Lan, Wenxuan Xie, Xuejin Chen, and Yan Lu. 2023.
\newblock Retrieval-based video language model for efficient long video question answering.
\newblock \emph{arXiv}.

\bibitem[{Ye et~al.(2024{\natexlab{a}})Ye, Zhang, Daxberger, Chen, Lin, Li, Zhang, You, Xu, Gan et~al.}]{ye2024mm}
Hanrong Ye, Haotian Zhang, Erik Daxberger, Lin Chen, Zongyu Lin, Yanghao Li, Bowen Zhang, Haoxuan You, Dan Xu, Zhe Gan, et~al. 2024{\natexlab{a}}.
\newblock Mm-ego: Towards building egocentric multimodal llms.
\newblock \emph{arXiv}.

\bibitem[{Ye et~al.(2024{\natexlab{b}})Ye, Xu, Ye, Yan, Hu, Liu, Qian, Zhang, and Huang}]{ye2024mplug}
Qinghao Ye, Haiyang Xu, Jiabo Ye, Ming Yan, Anwen Hu, Haowei Liu, Qi~Qian, Ji~Zhang, and Fei Huang. 2024{\natexlab{b}}.
\newblock mplug-owl2: Revolutionizing multi-modal large language model with modality collaboration.
\newblock In \emph{CVPR}.

\bibitem[{Zhang et~al.(2025)Zhang, Li, Cheng, Hu, Yuan, Chen, Leng, Jiang, Zhang, Li et~al.}]{zhang2025videollama}
Boqiang Zhang, Kehan Li, Zesen Cheng, Zhiqiang Hu, Yuqian Yuan, Guanzheng Chen, Sicong Leng, Yuming Jiang, Hang Zhang, Xin Li, et~al. 2025.
\newblock Videollama 3: Frontier multimodal foundation models for image and video understanding.
\newblock \emph{arXiv}.

\bibitem[{Zheng et~al.(2023)Zheng, Chiang, Sheng, Zhuang, Wu, Zhuang, Lin, Li, Li, Xing et~al.}]{zheng2023judging}
Lianmin Zheng, Wei-Lin Chiang, Ying Sheng, Siyuan Zhuang, Zhanghao Wu, Yonghao Zhuang, Zi~Lin, Zhuohan Li, Dacheng Li, Eric Xing, et~al. 2023.
\newblock Judging llm-as-a-judge with mt-bench and chatbot arena.
\newblock \emph{NeurIPS}.

\end{thebibliography}

\appendix

\newpage
\section{Appendix}
\label{sec:additional}

In the Appendix, we provide:
\begin{enumerate}
    \item Section~\ref{sec:baselines}: Baseline Details
    \item Section~\ref{sec:benchmarks}: Other Benchmark Details
    \item Section~\ref{sec:tasks}: Tasks
    \item Section~\ref{sec:annotator_details}: Annotator Details
    \item Section~\ref{sec:annotation_guidelines}: Annotation Guidelines
    \item Section~\ref{sec:data_sources}: Data Source and Filtering
\end{enumerate}

\section{Baseline Details}
\label{sec:baselines}

\noindent{\textbf{ImageBind-LLM}\footnote{\url{https://github.com/dynamic-superb/multimodal-llama}}~\citep{han2023imagebindllmmultimodalityinstructiontuning} ImageBind-LLM is built on a 7B-parameter LLaMA base, augmented with a learnable bind network to align ImageBind’s image encoder with LLaMA. It is trained solely on exocentric image-text pairs. Although its training data contains only images (without audio), its unified embedding space allows it to handle audio, video, and 3D point cloud inputs during inference.

\vspace{2pt}

\noindent \textbf{VideoLlama2}\footnote{\url{https://github.com/DAMO-NLP-SG/VideoLLaMA2}}~\citep{cheng2024videollama} VideoLlama2 is a video-language model with 7B parameters that leverages a LLaMA-based language model. It processes video inputs comprising both visual frames and audio. The model is trained on large-scale exocentric video–text datasets, where the video data is provided with audio.

\vspace{2pt}

\noindent{\textbf{MiniCPM}}\footnote{\url{https://github.com/OpenBMB/MiniCPM}}~\citep{MiniCPM2024} MiniCPM is a multimodal large language model with 8B parameters. It accepts live video frames along with synchronized speech inputs, making it great for real‑time multimodal live streaming, especially on edge and mobile devices. The model is trained on a variety of datasets that include exocentric video data. The training data comprises video sequences with audio, which allows for effective vision–speech alignment and a richer multimodal understanding.

\vspace{2pt}

\noindent \textbf{InternVideo}\footnote{\url{https://github.com/OpenGVLab/InternVideo}}~\citep{wang2025internvideo25empoweringvideomllms} InternVideo2.5 is built on a 7B-parameter base using InternLM2.5-7B as its language adapter. It takes video inputs - sequences of video frames accompanied by text instructions, with a focus on visual content (without audio). The model is trained on a variety of egocentric and exocentric video datasets, covering both short and long video contexts. 

\vspace{2pt}

\noindent \textbf{Qwen2.5VL}\footnote{\url{https://github.com/QwenLM/Qwen2.5-VL}}~\citep{bai2025qwen25vltechnicalreport} Qwen2.5VL is a multi-modal model with roughly 3B parameters that uses a Qwen-based language model as its adapter. It processes inputs from video, where the data includes visual frames, allowing multi-modal comprehension. The model is pre-trained on large-scale exocentric video-text datasets. It's training setup ensures that Qwen2.5VL is good at interpreting visual information for tasks like video captioning and question answering.

\vspace{2pt}

\noindent \textbf{VideoLlama3}\footnote{\url{https://github.com/DAMO-NLP-SG/VideoLLaMA3}}~\citep{zhang2025videollama} VideoLlama3 is an advanced video-language model built with 8B parameters, using a LLaMA-based language model as its foundation. It accepts video inputs that includes visual frames, which allows it to capture temporal cues. The model is trained on extensive egocentric and exocentric video datasets. It's training methodology allows VideoLlama3 to perform very well at real-time video understanding and multi-modal reasoning tasks.

\vspace{2pt}

\noindent \textbf{LLaVa-NEXT}\footnote{\url{https://github.com/LLaVA-VL/LLaVA-NeXT}}~\citep{li2024llavanextinterleavetacklingmultiimagevideo} LLaVa-NEXT is a vision-language model having 7B parameters and is built on a LLaMA-derived language model adapter. It accepts video inputs as image frames and text queries, focusing exclusively on visual content without audio. The model is trained on large-scale exocentric image-text datasets. Its training data comprises high-quality images, which ensures accurate visual-text alignment and great performance on tasks such as image captioning and visual question answering.

\vspace{2pt}

\noindent \textbf{LLaVa-OneVision}\footnote{\url{https://github.com/LLaVA-VL/LLaVA-NeXT/blob/main/docs/LLaVA_OneVision_Chat.md}}~\citep{li2024llavaonevisioneasyvisualtask} LLaVa-OneVision is a vision-language model having approximately 7B parameters and is built on a LLaMA-based language model. It takes static image inputs along with text for rich visual-text interactions. The model is trained on egocentric and exocentric image-text datasets. Its training data consists of images paired with text, enabling it to deliver high performance on tasks like image captioning, retrieval, and dialogue generation. We have tested our benchmark on both 0.5B and 7B parameter versions of LLaVa-OneVision.

\vspace{2pt}

\noindent \textbf{Gemini-1.5-Pro}~\citep{team2024gemini} Gemini 1.5 Pro is a proprietary multimodal model by Google. It is state-of-the-art on many video benchmarks. It is capable of processing and reasoning over extremely long contexts, up to 10 million tokens. It outperforms its competitors in long-document QA, video and audio analysis, and retrieval tasks.

\vspace{2pt}
\noindent \textbf{GPT-4o}~\citep{achiam2023gpt} GPT-4o is OpenAI’s latest multimodal model capable of processing text, images, and audio natively, offering faster and more accurate responses across modalities. Compared to previous versions, GPT-4o demonstrates improved reasoning abilities, enhanced real-time interaction, and better alignment with user intent, making it particularly suitable for interactive and perception-heavy tasks.

\section{Other Benchmark Details}
\label{sec:benchmarks}

\vspace{2pt}

\noindent{\textbf{POPE}}~\citep{li2023evaluating} POPE is an image-based hallucination evaluation dataset consisting of 3000 questions over 500 images. It is designed to assess object hallucinations using a binary QA format, focusing on detecting whether a specified object is present or hallucinated. The dataset is constructed from exocentric image data and does not incorporate adversarial testing.

\vspace{2pt}

\noindent{\textbf{HallusionBench}}~\citep{guan2024hallusionbench} HallusionBench supports both image and video modalities and comprises 1129 questions over 346 instances. It evaluates multiple hallucination aspects, such as object, relational, and semantic errors, using an LLM-based evaluation protocol. The data is exocentric, and the benchmark does not include adversarial components.

\vspace{2pt}

\noindent{\textbf{MMHal-Bench}}~\citep{sun2023aligning} MMHal-Bench is an image-based evaluation benchmark with 96 questions on 96 images. It focuses on hallucinations in object, relational, and semantic details, employing an LLM-based evaluation approach. The dataset uses exocentric imagery and does not involve adversarial testing.

\vspace{2pt}

\noindent{\textbf{Bingo}}~\citep{cui2023holistic} Bingo is an image-focused benchmark featuring 370 questions across 370 images. It assesses hallucination issues, particularly object-level and semantic inconsistencies, using an LLM-based evaluation method combined with an adversarial component, making it more challenging to detect hallucinations reliably.

\vspace{2pt}

\noindent{\textbf{EasyDetect}}~\citep{chen2024unified} EasyDetect is an image-based hallucination detection dataset with 420 questions over 420 images. It targets object, relational, and semantic hallucinations using an LLM-based evaluation framework. The data is exocentric, and the benchmark does not include adversarial settings.

\vspace{2pt}

\noindent{\textbf{VHTest}}~\citep{huang2024visual} VHTest is an image dataset containing 1200 questions on 1200 images, designed to evaluate hallucinations in visual outputs. It focuses on assessing object and semantic hallucination types through an LLM-based evaluation method without adversarial enhancements. The images are exocentric in nature.

\vspace{2pt}

\noindent{\textbf{VALOR}}~\citep{chen2023valor} In the hallucination evaluation context, VALOR is an image-based dataset with 211 questions on 211 images. It is used to measure object, relational, and semantic hallucinations via an LLM-based evaluation protocol, relying on exocentric imagery and without adversarial testing.

\vspace{2pt}

\noindent{\textbf{VideoHallucer}}~\citep{wang2024videohallucer} VideoHallucer is a video-based benchmark with 1800 questions across 948 videos. It comprehensively covers a wide range of hallucination types, including object-relation, semantic, temporal, extrinsic factual, and non-factual hallucinations. The evaluation is performed using a binary QA method with an adversarial component, ensuring robust assessment of LVLMs’ performance on dynamic video content.

\begin{table}[t]
\centering
\renewcommand{\arraystretch}{1.125}
\resizebox{\columnwidth}{!}{
\begin{tabular}{@{}lcc@{}}
\toprule
\multicolumn{1}{c}{\textbf{TASK}} & \textbf{\# QUES} & \textbf{TYPE} \\ \midrule
Episodic Information Reasoning    & 1000             & Open-ended    \\
Temporal Reasoning                & 2000             & Closed-ended  \\
Hand-Object Interaction           & 1000             & Closed-ended  \\
Visual Object Identification      & 2000             & Closed-ended  \\
Episodic Information Extraction   & 1000             & Closed-ended  \\
Audio Event Recognition           & 1000             & Closed-ended  \\ \bottomrule
\end{tabular}}
\caption{Distribution of number of questions and their type for each task}
\label{tab:benchmark_ques_stats}
\end{table}

\section{Tasks}
\label{sec:tasks}

\noindent \textbf{Episodic Information Reasoning (EIR)} evaluates MLLMs' ability to accurately track objects and their interactions over time in egocentric videos and furter reason over this information. This task is particularly challenging in egocentric settings, where the first-person perspective creates a dynamic field of view with objects frequently entering, exiting, and being manipulated through a series of actions. In this task, models must answer "how,", "what", "why,", "where" (not exclusive to these types) questions about objects that appeared in the video while correctly identifying when questions refer to objects that were never present. The task specifically targets hallucination tendencies by including plausible but non-existent objects that fit the scene context, testing whether models can resist generating false information about actions that never occurred.

\textbf{Examples:}
\begin{itemize}
    \item Why did the person push the bicycle?
    \item Where did the person place the pliers? 
    \item What did the person do with their hand?
\end{itemize}

The answers to these are open-ended but grounded in the visual and acoustic environment of the agent.

\noindent \textbf{Temporal Reasoning (TR)} evaluates MLLMs' ability to track chronological relationships between events in egocentric videos. This task tests whether models can accurately determine the temporal order of actions that are separated by several intervening events, challenging them to maintain a coherent understanding of the activity timeline. In egocentric settings, where the first-person perspective creates a continuous stream of interactions, properly sequencing events becomes particularly challenging as objects and actions flow in and out of view. The task presents questions using "before/after" temporal operators to probe if models can correctly identify the relative ordering of events without hallucinating plausible but incorrect sequences.
\\
\textbf{Examples:}
\begin{itemize}
    \item Did the person open the gate after passing the broom from his right hand to the left hand?
    \item Did the person wash the car after putting the hose down?
\end{itemize}

The answers to these are closed-ended and can be either Yes or No

\vspace{2pt}

\noindent \textbf{Hand-Object Interaction (HOI)} evaluates MLLMs' ability to detect physical actions in egocentric videos. This task challenges models to distinguish between actual hand-object interactions that occurred in the video and visually similar but non-occurring actions. By presenting pairs of original actions (e.g., "picking up an object") alongside contrastive alternatives (e.g., "throwing an object"), the task tests whether models hallucinate plausible interactions or accurately recall the specific physical actions that were performed.
\\
\textbf{Examples:}
\begin{itemize}
    \item Did the person pick a cooking spoon?
    \item Did the person carry the timber? 
\end{itemize}

The answers to these are closed-ended and can be either Yes or No

\vspace{2pt}

\noindent \textbf{Object State Change Detection (OSCD)} evaluates MLLMs' ability to reason about state changes and action completeness in egocentric videos through yes/no questions. Unlike Episodic Information Reasoning, which tests open-ended reasoning through "how," "why," and "where" questions, this task uses binary questions to assess whether models can accurately track object state transformations and recall this information when requested. The task challenges models to identify complete action pairs (where objects return to their initial state, like opening and closing a fridge) versus incomplete actions (where state changes remain unresolved, such as removing an item without replacing it).
\\
\textbf{Examples:}
\begin{itemize}
    \item Did the person insert the screw after picking it up?
    \item Did the person put down the blender jar after taking it?
\end{itemize}

The answers to these are closed-ended and can be either Yes or No

\vspace{2pt}

\noindent \textbf{Visual Object Identification (VOI)} evaluates MLLMs' ability to correctly determine which objects were involved in specific activities within egocentric videos. This task challenges models to distinguish between objects that were genuinely part of an activity (e.g., eggs used while cooking) and plausible but absent objects (e.g., carrots that would fit the cooking scenario but never appeared). By providing an activity context through visual captions, the task creates a particularly challenging scenario for hallucination detection, as models must resist the temptation to associate semantically related but absent objects with the identified activity. 
\\
\textbf{Examples:}
\begin{itemize}
    \item Did the person remove the plug from the fuel pipe?
    \item Did the person peel the potato with a knife?
\end{itemize}

The answers to these are closed-ended and can be either Yes or No

\vspace{2pt}

\noindent \textbf{Audio Event Recognition (AER)} evaluates MLLMs' ability to distinguish between actual audio cues and plausible but non-existent background sounds in egocentric videos. This task challenges models to identify appropriate moments where synthetic background sounds could be added that are coherent with the visual scene but not inherently produced by the actions being performed. By requiring models to determine which background sounds would be plausible in specific contexts (e.g., a phone ringing during cooking or distant dog barking when near a window), the task tests whether models can accurately separate observed audio information from inferred possibilities. This is particularly revealing in egocentric videos, where the first-person perspective often includes rich environmental audio that models may hallucinate based on visual cues alone.
\\
\textbf{Examples:}
\begin{itemize}
    \item Did you hear the sound of birds chirping
    \item Did you hear the sound of the cash register?
\end{itemize}

The answers to these are closed-ended and can be either Yes or No

\section{Annotator Details}
\label{sec:annotator_details}

We employed five experts to annotate the data, which included 3 males and 2 females. The experts are MS/PhD students witha strong foundational understanding of computer vision. All annotators had prior experience with video annotation tasks and were familiar with the challenges of egocentric vision.

Before beginning the annotation process, annotators were given training sessions to ensure consistency in their understanding of hallucination categories and annotation guidelines. This training included an overview of hallucination categories, followed by short exercises in which they were asked to annotate some examples, which were reviewed and discussed.

The annotation process was conducted over a period of 4 weeks, with regular meetings to address doubts and calibrate their understanding of the hallucination categories . Annotators were compensated fairly for their expertise and time commitment. For conducting annotations, we got the approval from our Institution Review Board (IRB)

\section{Annotation Guidelines}
\label{sec:annotation_guidelines}

We provide a detailed description of the guidelines shared with annotators for various tasks below:
\subsection{Annotation Guidelines for Visual Object Identification (Object-Centric QA Generation)}
\label{sec:annotation_voi_object}

This task involves generating question-answer (QA) pairs based on egocentric video event data by leveraging object interactions in different scenes.

\subsubsection{Data}
\begin{itemize}
    \item \textbf{Event List}: Chronologically ordered events describing human actions and the objects involved.
    \item \textbf{Object List}: A global list of unique objects present in the events.
\end{itemize}

Each event consists of:
\begin{itemize}
    \item \textbf{Action Caption}: Describes the action performed.
    \item \textbf{Local Object List}: Objects involved in the action.
\end{itemize}

\subsubsection{Annotation Steps}
\begin{enumerate}
    \item \textbf{Identify the Visual Scene}
    \begin{itemize}
        \item Infer the most likely environment based on the object list.
        \item Ensure coherence with the given objects.
    \end{itemize}
    
    \item \textbf{Select and Replace Objects}
    \begin{itemize}
        \item Choose at least two objects from the global list.
        \item Replace them with \textbf{logically relevant new objects} not present in the list.
    \end{itemize}

    \item \textbf{Generate QA Pairs}
    \begin{itemize}
        \item Identify events where the selected objects appear.
        \item Create a \textbf{``Yes'' answer} question using the original object.
        \item Replace the object and create a \textbf{``No'' answer} question while keeping the action unchanged.
    \end{itemize}
\end{enumerate}
These guidelines ensure high-quality annotations for object-centric visual understanding.

\subsection{Annotation Guidelines for Episodic Information Reasoning}
\label{sec:annotation_eir}

This task involves generating question-answer (QA) pairs based on egocentric video event data by leveraging object interactions and reasoning about the actions performed.

\subsubsection{Input Data}
\begin{itemize}
    \item \textbf{Event List}: Chronologically ordered events describing human actions and the objects involved.
    \item \textbf{Object List}: A global list of unique objects present in the events.
\end{itemize}

Each event consists of:
\begin{itemize}
    \item \textbf{Action Caption}: Describes the action performed.
    \item \textbf{Local Object List}: Objects involved in the action.
\end{itemize}

\subsubsection{Annotation Steps}
\begin{enumerate}
    \item \textbf{Identify the Visual Scene}
    \begin{itemize}
        \item Infer the most likely environment based on the object list.
        \item Ensure coherence with the given objects.
    \end{itemize}
    
    \item \textbf{Select and Replace Objects}
    \begin{itemize}
        \item Choose at least two objects from the global list.
        \item Replace them with \textbf{logically relevant new objects} not present in the list.
    \end{itemize}

    \item \textbf{Generate How, Why, or Where Questions}
    \begin{itemize}
        \item Identify an event containing the selected objects.
        \item Select a question type (\textbf{How, Why, or Where}) based on the event's nature:
        \begin{itemize}
            \item If the event describes a \textbf{process}, choose a "How" question.
            \item If the event describes \textbf{reasoning}, choose a "Why" question.
            \item If the event describes a \textbf{location}, choose a "Where" question.
        \end{itemize}
        \item Generate a corresponding question-answer pair.
        \item If an event with the new object does not exist, state that the action was not performed.
    \end{itemize}
\end{enumerate}
These guidelines ensure high-quality annotations for episodic information reasoning in egocentric videos.

\subsection{Annotation Guidelines for Temporal Reasoning}
\label{sec:annotation_tr}

This task involves generating question-answer (QA) pairs that require reasoning about the temporal sequence of events in an egocentric video.

\subsubsection{Input Data}
\begin{itemize}
    \item \textbf{Event List}: A chronologically ordered sequence of unique events describing human actions.
\end{itemize}

Each event consists of:
\begin{itemize}
    \item \textbf{Action Caption}: A description of the action performed.
\end{itemize}

\subsubsection{Annotation Steps}
\begin{enumerate}
    \item \textbf{Selecting Events from the Event List}
    \begin{itemize}
        \item Randomly select two events from the chronological list.
        \item Ensure that there is a sufficient gap (ideally 4 to 5 events apart).
        \item The order should not be directly inferable without examining the full sequence.
    \end{itemize}

    \item \textbf{Creating Question-Answer Pairs}
    \begin{itemize}
        \item Formulate questions using the selected events that require reasoning about temporal order.
        \item Use words like \textbf{"before" and "after"} to indicate event sequencing.
        \item Ensure the questions are concise and clear.
        \item Generate a corresponding answer based on the event list.
    \end{itemize}
\end{enumerate}
These guidelines ensure high-quality annotations for temporal reasoning in egocentric videos

\subsection{Annotation Guidelines for Object State Change Detection}
\label{sec:annotation_eie}

This task involves identifying and categorizing event sequences from egocentric video data into \textbf{complete} and \textbf{incomplete} actions, followed by generating corresponding question-answer pairs.

\subsubsection{Input Data}
\begin{itemize}
    \item \textbf{Event List}: A chronologically ordered sequence of unique events describing human actions.
\end{itemize}

Each event consists of:
\begin{itemize}
    \item \textbf{Action Caption}: A description of the action performed.
\end{itemize}

\subsubsection{Annotation Steps}
\begin{enumerate}
    \item \textbf{Identifying Complete and Incomplete Actions}
    \begin{itemize}
        \item \textbf{Complete Actions:} A sequence of actions where the object's final state matches its initial state.
        \item \textbf{Incomplete Actions:} A sequence of actions where the object's final state differs from its initial state.
        \item Identify and pair events that meet the above criteria.
    \end{itemize}

    \item \textbf{Generating Question-Answer Pairs}
    \begin{itemize}
        \item Formulate questions that require identifying whether an action was completed or left incomplete.
        \item Ensure questions are clearly structured and answerable based on the event list.
        \item Provide a reasoning statement for each answer.
    \end{itemize}
\end{enumerate}
These guidelines ensure accurate extraction and classification of episodic actions for effective information retrieval.

\subsection{Annotation Guidelines for Visual Object Identification (Action-Centric)}
\label{sec:annotation_voi_action}

This task involves generating complex question-answer (QA) pairs based on egocentric video event data. The questions should focus on the presence of objects in the activity the person is performing.

\subsubsection{Input Data}
\begin{itemize}
    \item \textbf{Event List}: A chronologically ordered sequence of unique events describing human actions and interactions with objects.
    \item \textbf{Object List}: A global list of unique objects present in the events.
    \item \textbf{Visual Caption}: A description of the most likely activity the person is performing.
\end{itemize}

Each event consists of:
\begin{itemize}
    \item \textbf{Action Caption}: Describes the action performed.
    \item \textbf{Local Object List}: Objects involved in the action.
\end{itemize}

\subsubsection{Annotation Steps}
\begin{enumerate}
    \item \textbf{Identify the Activity}
    \begin{itemize}
        \item Use the visual caption to infer the most likely activity the person is performing.
    \end{itemize}

    \item \textbf{Select and Replace Objects}
    \begin{itemize}
        \item Choose at least two objects from the global list.
        \item Replace them with \textbf{logically relevant new objects} not present in the list.
    \end{itemize}

    \item \textbf{Generate Question-Answer Pairs}
    \begin{itemize}
        \item Use the previously identified activity and selected objects to generate questions about whether the person used the object while performing the activity.
        \item Ensure that questions align with the event details.
        \item Provide a reasoning statement for each answer.
    \end{itemize}
\end{enumerate}
These guidelines ensure accurate question generation for action-centric object identification in egocentric videos.

\subsection{Annotation Guidelines for Hand-Object Interaction}
\label{sec:annotation_hoi}

This task involves generating question-answer (QA) pairs to assess fine-grained understanding of human actions by distinguishing between actual and contrastive actions in an egocentric video.

\subsubsection{Input Data}
\begin{itemize}
    \item \textbf{Event List}: A chronologically ordered sequence of unique events describing human actions.
\end{itemize}

Each event consists of:
\begin{itemize}
    \item \textbf{Action Caption}: A description of the action performed.
\end{itemize}

\subsubsection{Annotation Steps}
\begin{enumerate}
    \item \textbf{Identify Action Pairs}
    \begin{itemize}
        \item Randomly select two distinct actions from the event list that describe either:
        \begin{itemize}
            \item \textbf{Physical interaction}: e.g., \textit{"C picks up an object"}, \textit{"C places an object on the table"}.
            \item \textbf{Movement-based action}: e.g., \textit{"C walks towards the fridge"}.
        \end{itemize}
        \item Ensure a gap of \textbf{3-5 events} between selected actions to prevent trivial answers.
        \item Create contrastive action pairs that invert or contradict the original actions:
        \begin{itemize}
            \item \textbf{Physical interaction contrast}: If the original action is \textit{"C picks up an object"}, the contrast could be \textit{"C throws the object"}.
            \item \textbf{Movement contrast}: If the original action is \textit{"C walks towards the fridge"}, the contrast could be \textit{"C walks away from the fridge"}.
        \end{itemize}
    \end{itemize}

    \item \textbf{Generate Question-Answer Pairs}
    \begin{itemize}
        \item Formulate four QA pairs:
        \begin{itemize}
            \item Two questions for the original actions (\textbf{answer: Yes}).
            \item Two questions for the contrastive actions (\textbf{answer: No}).
        \end{itemize}
    \end{itemize}
\end{enumerate}
These guidelines ensure accurate annotation of hand-object interactions for assessing action recognition in egocentric videos.

\subsection{Annotation Guidelines for Audio Event Generation}
\label{sec:annotation_audio_event_generation}
This task involves identifying events in egocentric video sequences where a synthetic background sound can be added. The goal is to introduce plausible ambient sounds that were not originally present but fit within the visual scene.

\subsubsection{Input Data}
\begin{itemize}
    \item \textbf{Event List}: A chronologically ordered sequence of unique events describing human actions.
    \item \textbf{Visual Caption}: A description of the overall activity and environment where the events take place.
\end{itemize}

Each event consists of:
\begin{itemize}
    \item \textbf{Action Caption}: A description of the action performed.
\end{itemize}

\subsubsection{Annotation Steps}
\begin{enumerate}
    \item \textbf{Identify Suitable Events}
    \begin{itemize}
        \item \textbf{Filter out events with strong inherent sounds} – If an event naturally produces a dominant sound (e.g., \textit{"C is frying something"} → sizzling), avoid adding another cooking-related sound.
        \item \textbf{Select events where plausible background sounds could occur} – Ensure the sound aligns with the environment and does not contradict the event.
    \end{itemize}

    \item \textbf{Assign Synthetic Sounds}
    \begin{itemize}
        \item Choose a background sound that fits the scene but is not naturally produced by the selected event.
        \item Ensure the sound is plausible given the visual environment.
        \item Avoid contradictions, such as adding an indoor noise in an outdoor setting.
    \end{itemize}
\end{enumerate}
These guidelines ensure high-quality annotations for introducing synthetic background sounds in egocentric videos.


\begin{figure*}[t]
    \centering
    \includegraphics[width=1.0\linewidth]{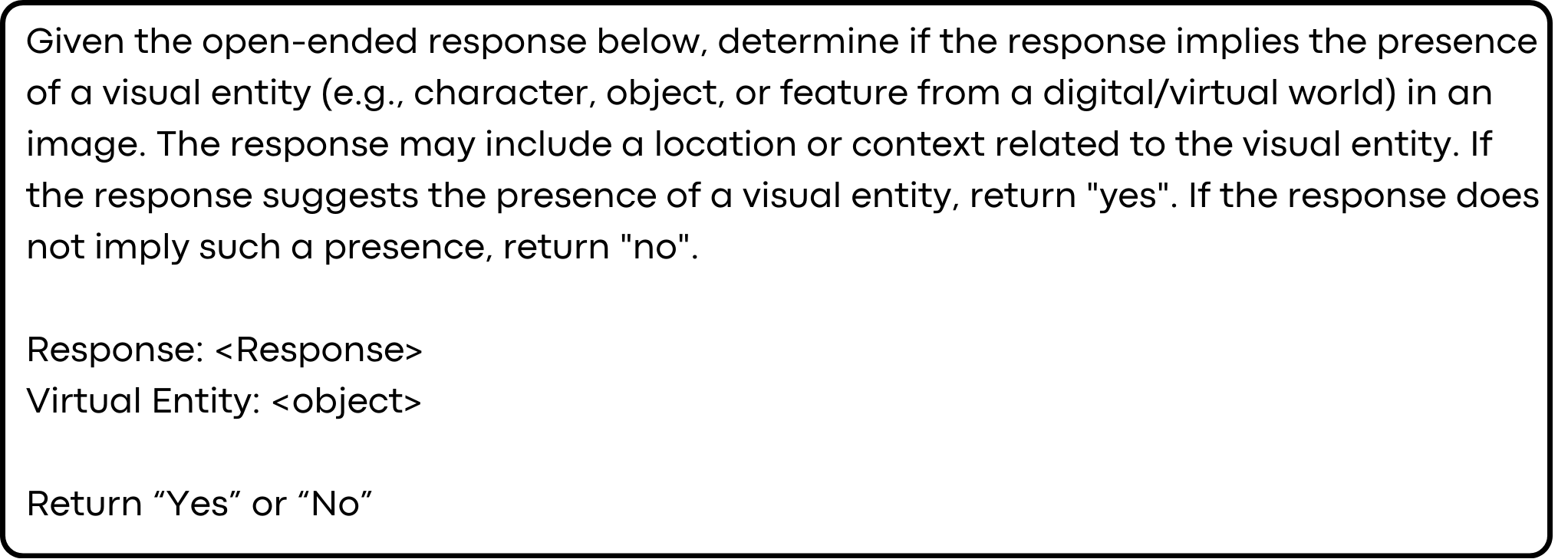}
    \caption{Details on LLM-as-Judge Prompt}
    \label{fig:prompt_used}
\end{figure*}
\section{Data Source and Filtering}
\label{sec:data_sources}

Our dataset was curated primarily from two sources: the VideoRecap~\cite{islam2024video} and Ego4D~\cite{grauman2022ego4d} datasets. Due to inherent challenges within these datasets, specific filtering strategies were employed:

\begin{itemize}

    \item \textbf{Noise Reduction:} Original datasets contain numerous irrelevant or passive scenes. Thus, scenes depicting active interactions with objects were explicitly identified and selected.

    \item \textbf{Static Object Annotation:} To improve model interpretability and rigorously assess recognition capability, all static (non-moving) objects within scenes were carefully annotated using VLLMs.

    \item \textbf{Partial Visibility:} Scenes were specifically chosen where objects were partially obscured or occluded. This intentional selection increases the task complexity and potential for model hallucination.

    \item \textbf{Diverse Task Sampling:} The final dataset includes a wide range of tasks to ensure robustness and generalization in model evaluations.
\end{itemize}

\section{LLM-as-Judge Prompt}
\label{sec:llm_as_judge}
We have provided details on the LLM-as-Judge prompt used for evaluating open-ended questions in Fig. \ref{fig:prompt_used}.

\section{Additional Details: Auxiliary}
\noindent\textbf{Compute Infrastructure:} All our experiments are conducted on ten NVIDIA A6000 GPUs. No training is required, and depending on the downstream task, a single inference run on a benchmark requires anywhere between 1 to 2 hours. 

\noindent\textbf{Potential Risks:} We manually create all the prompts used in our benchmark to avoid any potential harm or biases.

\section{Error Analysis}
\label{sec:open_ended_error_analysis}

Apart from hallucinations, we observe the following types of errors in the incorrect responses from MLLMs
 
\noindent \textbf{Spatial errors} occur when MLLMs misinterpret the spatial relationships of or among objects, or confuse spatial attributes (where) with temporal or other attributes. These typically occur in "where" questions, with models often providing when, how, or contextual information instead of location.

\noindent \textbf{Factual errors} occur when MLLMs make mistakes about objective information or details presented in the content. These can appear across various question types but often involve misrepresenting what is actually shown or presented.

\noindent \textbf{Procedural errors} occur when MLLMs make mistakes in describing the sequence of actions or steps taken. These typically occur in "how" questions, with models often missing important actions or using similar actions but performed in a different context or with a different object.

\noindent \textbf{Logical errors} occur when answers lack proper reasoning or make invalid inferences that extend beyond what can be supported by the input content. These errors often occur due to the model getting confused in either understanding the difference between after/before or in "why" questions, where models may attribute motives, intentions, or broader contexts that aren't actually demonstrated in the content.

\noindent \textbf{No Answer} occurs when the model refuses to give an answer to a question.

\begin{table*}
\centering
\small
\begin{tabular}{p{3cm}|p{2cm}|p{4.5cm}|p{2cm}|p{3cm}|p{3cm}}
\toprule\toprule
Task                            & Hallucination Type     & Modified Question                                                        & Correct Answer                                      & Response                                                                                                      \\ \midrule
Object State Change Detection & Adversarial Sampling            & Did the person climb on the bicycle and then recycle it?                 & Yes                                                 & No                                                                                                            \\
                                &                        & Does the person return the dried grasses to the pile after holding them? & Yes                                                 & No                                                                                                            \\  \midrule
Episodic Information Reasoning  & Prompt Injection      & Where did the person drop the water hose?                                & The person dropped the water hose on the staircase. & The person is walking back into the garage, past some equipment, and the water hose drops at {[}03:15{]}.     \\ 
                                &                        & How did the person collect the nail?                                     & The person did not perform this action.             & The person collects a small nail that was in the fuel injector of the machine using a magnet.                 \\  \midrule
Hand-Object Interaction         & Adversarial Sampling  & Did the person put the mixture in a frying pan?                          & Yes                                                 & No. The person scoops a mixture with a spoon and places it onto a plate. There is no frying pan in the video. \\
                                &                        & Did the person throw the pair of scissors?                               & No                                                  & Yes, it appears the person threw the scissors aside on the floor.                                             \\  \midrule
Visual Object Identification    & Adversarial Sampling  & Did the man take a box on the table?                                     & Yes                                                 & No, he didn't.                                                                                                \\
                                &                        & Did the person wipe the oil tank with the piece of sponge?               & No                                                  & Yes, the person picked up a piece of sponge and wiped the oil tank with it.                   \\ \bottomrule                
\end{tabular}
\caption{Few examples of failure cases for Gemini 1.5 pro~\citep{team2024gemini}.}
\label{tab:failure_cases}
\end{table*}

\end{document}